\documentclass[10pt,journal,compsoc]{IEEEtran}
\usepackage{latexsym,amssymb,amsmath,amsbsy,amsopn,amstext,amsthm,amsxtra,color,bm,calc,ifpdf}
\usepackage{float}
\usepackage{enumerate}
\usepackage{fancyhdr}
\usepackage{listings}
\usepackage{multicol}
\usepackage{multirow}
\usepackage{makeidx}
\usepackage{makecell}
\usepackage{xcolor}

\ifCLASSOPTIONcompsoc
  \usepackage[nocompress]{cite}
\else
  \usepackage{cite}
\fi

\usepackage{tabularx,booktabs}

\ifCLASSINFOpdf
  \usepackage[pdftex]{graphicx}
  \graphicspath{{figures/}}
  \DeclareGraphicsExtensions{.pdf,.jpeg,.png}
\else
  \usepackage[dvips]{graphicx}
  \graphicspath{{figures/}}
  \DeclareGraphicsExtensions{.eps}
\fi

\ifCLASSOPTIONcompsoc 
  \usepackage[caption=false,font=normalsize,labelfont=sf,textfont=sf]{subfig}
\else
  \usepackage[caption=false,font=footnotesize]{subfig}
\fi

\usepackage{algorithm}
\usepackage{algorithmicx}
\usepackage{algpseudocode}

\usepackage{array}

\usepackage{url}

\newcommand {\Eqref}[1]{Eq. (\ref{#1})}
\newcommand {\Figref}[1]{Fig. \ref{#1}}
\newcommand {\Romnum}[1]{\uppercase\expandafter{\romannumeral #1}}

\makeatletter
\let\NAT@parse\undefined
\makeatother
\usepackage{xr}
\usepackage{hyperref}  

\begin{document}
\title{Regularized Multi-output Gaussian Convolution Process with Domain Adaptation}

\author{Xinming~Wang,~\IEEEmembership{} Chao~Wang, ~\IEEEmembership{}  Xuan~Song, ~\IEEEmembership{}  Levi~Kirby, ~\IEEEmembership{}  Jianguo~Wu~\IEEEmembership{} 
\IEEEcompsocitemizethanks{\IEEEcompsocthanksitem X. Wang and J. Wu are with the Department of Industrial Engineering and Management, Peking University, Beijing 100089, China.\protect\\
E-mail: wang-xm20@stu.pku.edu.cn, j.wu@pku.edu.cn.}
\IEEEcompsocitemizethanks{\IEEEcompsocthanksitem C. Wang, X. Song and L. Kirby are with the Department of Industrial and Systems Engineering, the University of Iowa, Iowa City 52242, America.\protect\\
E-mail: \{chao-wang-2,  xuan-song,  levi-kirby\}@uiowa.edu.}
\thanks{(Corresponding authors: Chao Wang and Jianguo Wu)}
}

\markboth{IEEE TRANSACTIONS ON PATTERN ANALYSIS AND MACHINE INTELLIGENCE
}%
{Shell \MakeLowercase{\textit{et al.}}: Regularized Multi-output Gaussian Convolution Process with Domain Adaptation}

\IEEEtitleabstractindextext{%
\begin{abstract}
Multi-output Gaussian process (MGP) has been attracting increasing attention as a transfer learning method to model
multiple outputs. Despite its high flexibility and generality, MGP still faces two critical challenges when applied to transfer
learning. The first one is negative transfer, which occurs when there exists no shared information among the outputs.
The second challenge is the input domain inconsistency, which is commonly studied in transfer learning yet not explored in MGP. In this paper, we propose a regularized MGP modeling framework with domain adaptation to overcome these challenges. More
specifically, a sparse covariance matrix of MGP is proposed by using convolution process, where penalization terms are added to
adaptively select the most informative outputs for knowledge transfer. To deal with the domain inconsistency, a domain adaptation
method is proposed by marginalizing inconsistent features and expanding missing features to align the input domains among different
outputs. Statistical properties of the proposed method are provided to guarantee the performance practically and asymptotically. The
proposed framework outperforms state-of-the-art benchmarks in comprehensive simulation studies and one real case study of a
ceramic manufacturing process. The results demonstrate the effectiveness of our method in dealing with both the negative transfer and the domain inconsistency.
\end{abstract}

\begin{IEEEkeywords}
Gaussian process, transfer learning, convolution process, domain adaptation.
\end{IEEEkeywords}}

\maketitle

\IEEEdisplaynontitleabstractindextext
\IEEEpeerreviewmaketitle

\ifCLASSOPTIONcompsoc
\IEEEraisesectionheading{\section{Introduction}\label{sec:1-introduction}}
\else
\section{Introduction}
\label{sec:1-introduction}
\fi

\IEEEPARstart{G}{aussian} process regression (GPR) model has been gaining widespread applications in many fields,  e.g., computer experiments, geostatistics, and robot inverse dynamics \cite{Williams2006, Shi2011}.
As a powerful nonparametric method, it possesses many desirable and important properties, including excellent fitting capability for various functional relationships under some regularity conditions, providing not only predictions but also uncertainty quantification, and more importantly having closed-form expressions for both tasks.

The conventional GPR models are designed for single-output cases, i.e. the output is a scalar, which have been extensively studied in various applications \cite{Williams1996, Stegle2008}. 
Recently, there has been a growing interest in extending GPR models to multiple outputs, which are ubiquitous nowadays. 
A straightforward way to deal with multiple outputs is known as multi-kriging, which constructs models for each output independently \cite{Boyle2005}. 
It is clear that the multi-kriging impairs the modeling of covariance among outputs, especially when there is strong evidence for the existence of such relationship resulting from physics or constraints. 
Hence, multi-output Gaussian process, which can model correlation among outputs, has been attracting more attention as a joint prediction model. 
The study of MGP begins in geostatistics community known as Co-Kriging in the past few decades \cite{Haas1996}. 
In today's machine learning society, it is usually known as a multi-task learning method \cite{Pan2009}, which aims to learn all tasks/outputs simultaneously to achieve better model generalization. 
For example, in the anomaly detection for a manufacturing process, the joint modeling of multiple closely related sensor signals using MGP can help detect the anomaly in each signal more efficiently \cite{Cho2019}.

However, despite its wide applications, MGP-based multi-task learning requires that the data among these outputs are balanced, which might not be the case in practice. 
For example, in \cite{Cho2019},  jointly modeling two correlated pressure signals may not improve the anomaly detection efficiency for both signals if the samples from one signal are much less than the other. 
Combining MGP and transfer learning is an effective way of handling such problems. 
When the observed data for one output is rare or expensive to collect, it is needed to exploit useful information from other outputs whose data are abundant. 
In this work, we focus on making predictions for one output which is denoted as \emph{target}, by leveraging data in some related outputs which are denoted as \emph{sources}. 
For instance, in the robot inverse dynamics problem, the target is the torque at a joint when the robot is working with a new load, and the sources are the torques at the same joint when the robot works with other loads \cite{Chai2008}. 
The key to the MGP-based transfer learning is to extract and represent the underlying similarity among outputs and leverage information from source outputs to target output so as to improve the prediction accuracy \cite{Liu2018}. 
Specifically, this information transfer in MGP is achieved by constructing a positive semi-definite covariance matrix describing the correlation of data within and across the outputs \cite{Conti2010}. 
There are two categories of models for the covariance structure: separable models and non-separable models. 
Separable models are most widely used approaches, including intrinsic coregionalization model (ICM) \cite{Goovaerts1997}, linear model of coregionalization (LMC) \cite{Goulard1992}, and their extensions. 
These models use the Kronecker products of a coregionalization matrix and a covariance matrix of single GP to represent the covariance matrix of MGP. 
It is clear that the separable models are not suitable for transfer learning since it restricts the same covariance structure (from single GP) for both the sources and the target. 
On the other hand, the non-separable models overcome this limitation by using convolution process (CP) to construct the MGP and its covariance structure.
They build non-separable covariance function through a convolution operation and allow modeling each output with individual hyperparameters \cite{Majumdar2007}. 
This property makes them more flexible and superior to the separable models \cite{Alvarez2011}. 

Nevertheless, there are two critical issues to be considered when apply non-separable MGP to transfer learning. 
The first issue is  negative transfer, which occurs when the assumption of 'existence of shared information' breaks, i.e.,  learning source outputs will have negative impacts on the learning of target output. \cite{Rosenstein2005}. 
The root cause of this issue is the excessive inclusion of data into the learning process, which is an increasingly severe issue in the big data environment. In such conditions, only a portion of the source data is correlated with the target data  and it is desired to select the sources that yield the best transfer \cite{Weiss2016}. 
A recent work \cite{Kontar2020} proposed a two-stage strategy to alleviate the negative transfer in MGP. 
In this method, the first step is to train a two-output Gaussian process model between each source and the target. 
In the second step, the inverse of predictive standard deviation of each two-output model is adopted as the index to evaluate the negative transfer of each source and integrate the results of transfer learning.
However, the two-stage approach raises significant concerns of losing global information as it only measures the pairwise transferability. 
\cite{Kontar2018} establishes a mixed-effect MGP model which has the ability to infer the behavior of the target output when it is highly similar with some sources. 
However, this method cannot guarantee the optimal selection of related sources, which will be demonstrated in our case study.

The second critical issue is that the input domains of the source processes might be inconsistent with that of the target process. 
For example, in multilingual text categorization, data in different languages have different features and we can't directly combine them to train a classifier for the target data \cite{Li2013}. 
Another example is shown in our case study, where the goal is to conduct transfer learning for predicting product density between dry pressing process and additive manufacturing process. 
It is clear that two different manufacturing processes will have different process parameters (inputs) that contribute to the product density, e.g., the dry pressing process is dominated by temperature and pressure while the additive manufacturing process is influenced by solids loading  percent and temperature \cite{He2019}. 
The two processes share one common process parameter (temperature), yet they also have distinct process parameters, which makes the transfer learning of product density (output) a non-trivial task. 
Indeed, the input domain inconsistency is a common issue in transfer learning, and domain adaptation is usually used to overcome this issue. The basic idea of domain adaptation is to align the domains between source and target by transforming data into certain feature domain, and it mainly applies to classification methods such as logistic regression and support vector machine (SVM)  \cite{Daume2009, Shi2010, Li2013, Xiao2015}. 
These domain adaptation methods aim to find feature mapping by minimizing sum of the training loss of learner and the difference among inconsistent domains, through solving a convex optimization problem. 
However, the training loss of MGP is the negative log-likelihood function, which is a strongly non-convex function. 
Therefore, a unique estimation of the parameters in feature mapping cannot be guaranteed.
More importantly, applying the existing domain adaptation methods directly to MGP might fail the transfer learning due to the existence of negative transfer, i.e., minimizing the difference of features between a negative source and the target will aggravate the severity of negative transfer. To the best of our knowledge, there is no research simultaneously handling issues of domain inconsistency and negative transfer in the context of MGP.

To overcome the above challenges, we propose a comprehensive regularized multi-output Gaussian convolution process (MGCP) modeling framework. In some literature \cite{Kasarla2021}, MGP with CP-based covariance is also named as multi-output convolution Gaussian process (MCGP).
Our method focuses on mitigating negative transfer of knowledge while at the same time adapting inconsistent input domain. 
In this work, we assume that there is at least one shared input feature between the sources and the target. 
This assumption is also necessary to facilitate the transferability, i.e., there is nothing to transfer if all the inputs in the sources and the target are different. 
Instead of learning all outputs equivalently, the proposed framework is based on a special CP structure that emphasizes the knowledge transfer from all source outputs to the target output,  which features the unique characteristic of transfer learning and differentiates it from many existing multi-task learning methods. 
The computation complexity is also significantly  reduced from $O((qn+n_t)^3)$ to $O(qn^3+n_t^3)$ when modeling $q$ sources with $n$ data points in each and one target with $n_t$ data points due to this special CP structure. 
The major contributions of this work include:
\begin{enumerate}
\item 
Building upon this special CP structure, a global regularization framework is proposed, which can penalize un-correlated source outputs so that the selection of informative source outputs and transfer learning can be conducted simultaneously.
\item
We provide some theoretical guarantees for our method, including the connection between penalizing parameters and selecting source outputs, and the asymptotic properties of the proposed framework.
\item 
We propose to marginalize extra input features and expand missing input features in the source to align with the input domain of the target, so that the domain inconsistency can be solved. 
\end{enumerate}
Both the simulation studies and real case study demonstrate the effectiveness of our framework in selecting informative sources and transferring positive information even when the target is not quite similar to all the sources.

The remainder of this article is organized as follows. 
The general multi-output Gaussian process and convolution process modeling framework are stated in Section \ref{sec:2-preliminaries}. 
In Section \ref{sec:3-model development}, a detailed description of our regularized MGCP modeling framework for transfer learning is presented, including some statistical properties and domain adaptation technique. 
Section \ref{sec:4-numerical studies} presents numerical studies to show the superiority of the proposed method using both simulated data and real manufacturing data. 
The conclusion is given in Section \ref{sec:5-conclusion}. Technical proofs are relegated to the appendix. 

\section{Preliminaries}
\label{sec:2-preliminaries}

\color{black}
\subsection{Related works on MGCP}
As mentioned above, several multi-output Gaussian convolutional process has been investigated for multi-task learning recently.  To handle different kinds of outputs, e.g., continuous output and categorical output, \cite{Moreno2018} proposes a heterogeneous multi-output Gaussian process and conduct variational inference in training and forecasting. Considering that each output may have its unique feature which is not shared with other outputs, \cite{Kasarla2021} constructs a MGCP model, where each output consists of two parts: one part is correlated with other output, while the remaining part is independent of others. Compared with these works, our method tackles the problem of inconsistent input domain rather than heterogeneous outputs, and focuses on selecting informative sources in one output (target) prediction. 

Besides for multi-task learning, there are two works using MGCP for information transfer to one output \cite{Kontar2018, Kontar2020}. Both of them  pay no attention to the problem of  inconsistent input domain, which limits the available source data for them. In addition, negative transfer is not explored in \cite{Kontar2018}, and the two-strategy method in \cite{Kontar2020} only realizes sub-optimal performance in reducing negative transfer. More detailed comparison can be found in Section 4.5.

Computational load is a severe limitation for multi-output Gaussian process when dealing with large amount of data. In addition to the popular sparse approximation method using inducing variables \cite{Alvarez2011,Moreno2018}, \cite{Bruinsma2020} assumes that all $q$ outputs lie in a low-dimensional linear subspace, which can be represented by $\tilde{q} \ll q$ orthogonal basis process, and the computational complexity can be reduced to $O(\tilde{q} n^3)$. \cite{Yu2021} proposes an approach based on local GP experts, which partitions the input and output space into segments to train local experts, then combines them to form a model on full space. These techniques can be applied to our method when extending it to a big data environment. In this paper, we focus more on the effectiveness of our method in reducing negative transfer and handling inconsistent inputs.

Furthermore, convolutional-kernel-based Gaussian process has been applied to high-dimensional and structural data, e.g., image, graph and point cloud data \cite{van2017, Walker2019}. 
In these works, discrete convolution operation is applied on patch of pixels to construct covariance between two data samples. This type of Gaussian process can be applied to image or 3D mesh classification. However, in most of MGCP methods, including our proposed, the convolution operation  is continuous and applied on latent processes.

\color{black}
\subsection{Multi-output Gaussian Process}
\label{sec:2-mgp}
In this subsection, we will review some basic theories of Multi-output Gaussian process. 
Consider a set of $q$ source outputs ${f}_i : \mathcal{X} \mapsto \mathbb{R},\ i=1,...,q$ and one target output ${f}_t : \mathcal{X} \mapsto \mathbb{R}$, where $\mathcal{X}$ is an input domain applied to all outputs. 
The $q+1$ outputs jointly follow some multi-output Gaussian process as
\begin{align}
(f_1,f_2,...,f_q,f_t)^T \sim \mathcal{GP}\left(\bm{0}, \mathcal{K}(\bm{x},\bm{x}^{\prime})\right),
\end{align}
where the covariance matrix $\mathcal{K}(\bm{x},\bm{x}^{\prime})$ is defined as 
\begin{align}
\left\{ \mathcal{K}(\bm{x},\bm{x}^{\prime}) \right\}_{ij}={\rm cov}_{ij}^f (\bm{x},\bm{x}^{\prime})={\rm cov} \left( f_i(\bm{x}), f_j(\bm{x}^{\prime}) \right),
\end{align}
$i,j \in \mathcal{I}= \{1,2,...,q,t\}$ and $\bm{x},\bm{x}^{\prime} \in \mathcal{X}$. Let $\mathcal{I}^S=\{1,2,...,q\}$ denote the index set of source outputs. The element $\left\{ \mathcal{K}(\bm{x},\bm{x}^{\prime}) \right\}_{ij}$ corresponds to the dependency between $f_i(\bm{x})$ and $f_j(\bm{x}^{\prime})$. 

Assume that the observation at point $\bm{x}$ is
\begin{align}
y_i(\bm{x})=f_i(\bm{x})+\epsilon_i, i \in \mathcal{I},
\label{eq:decomposition}
\end{align}
where $\epsilon_i \sim \mathcal{N}(0,\sigma_i^2)$ is independent and identically distributed (i.i.d) Gaussian noise assigned to the $i$th output.
Denote the observed data for the $i$th output as $\mathcal{D}_i=\left\{\bm{X}_i, \bm{y}_i\right\}$, where $\bm{X}_i=(\bm{x}_{i,1},...,\bm{x}_{i,n_i})$, $\bm{y}_i=(y_{i,1},...,y_{i,n_i})^T$ are the collections of input points and associated observations,  and $n_i$ is the number of observations for the $i$th output.  
Suppose that $N=\sum_{i \in \mathcal{I}}n_i$.  
Let $\mathcal{D}^S=\left\{\mathcal{D}_i | i\in \mathcal{I}^S\right\}$ denote the observed data of $q$ source outputs and $\mathcal{D}=\left\{\mathcal{D}^S,\mathcal{D}_t\right\}$ denote all data.
 Define the matrix $\bm{X}$ and vector $\bm{y}$ for all input points and observations as
$\bm{X}=\left(\bm{X}_1,\bm{X}_2,...,\bm{X}_q,\bm{X}_t \right)$, 
$\bm{y}=\left(\bm{y}_1^T,\bm{y}_2^T,...,\bm{y}_q^T,\bm{y}_t^T \right)^T$.

Since GP is a stochastic process wherein any finite number of random variables have a joint Gaussian distribution, for any new input point $\bm{x}_*$ associated with the target output $f_t$, the joint distribution of all observations $\bm{y}$ and the target function value $f_t^*=f_t(\bm{x}_*)$ is
\begin{align}
\begin{pmatrix} \bm{y} \\ f_t^* \end{pmatrix} \sim \mathcal{N}
\begin{pmatrix}
	\begin{bmatrix} \bm{0} \\ 0 
	\end{bmatrix},
	\begin{bmatrix} \bm{K}(\bm{X},\bm{X})+\bm{\Sigma} & \bm{K} (\bm{X}, \bm{x}_*)\\
					\bm{K} (\bm{X}, \bm{x}_*)^T & {\rm cov}_{tt}^f(\bm{x}_*, \bm{x}_*)
	\end{bmatrix}
\end{pmatrix},
\label{eq:joint distribution}
\end{align}
where $\bm{K}(\bm{X},\bm{X}) \in \mathbb {R}^{N \times N}$ is a block partitioned covariance matrix whose $i,j$th block, $\bm{K}_{i,j}\in \mathbb{R}^{n_i \times n_j}$, represents the covariance matrix between the output $i$ and output $j$; $\bm{\Sigma}$ is a block diagonal noise covariance matrix with $\bm{\Sigma}_{i,j}=\sigma_i^2\bm{I}_{n_i}$ if $i=j$ and $\bm{0}$ otherwise; $\bm{K} \left(\bm{X}, \bm{x}_*\right)=\left( \bm{K}_{1,*}^T, \bm{K}_{2,*}^T,..., \bm{K}_{q,*}^T,\bm{K}_{t,*}^T \right)^T$ and $\bm{K}_{i,*}=\left( {\rm cov}_{i,t}^f (\bm{x}_{i,1},\bm{x}_*),{\rm cov}_{i,t}^f (\bm{x}_{i,2},\bm{x}_*),...,{\rm cov}_{i,t}^f (\bm{x}_{i,n_i},\bm{x}_*) \right)^T$. 
To simplify the notations, we introduce a compact form that $\bm{K}=\bm{K}\left(\bm{X},\bm{X}\right)$, $\bm{K}_*=\bm{K} (\bm{X}, \bm{x}_*)$ and $\bm{C}=\bm{K}+\bm{\Sigma}$.

Based on the multivariate normal theory, the posterior distribution of $f_t(\bm{x_*})$ given data $\{\bm{X},\bm{y}\}$ can be derived as
\begin{align}
f_t(\bm{x}_*)| \bm{X},\bm{y} \sim \mathcal{N}\left( \mu(\bm{x}_*), {V}_f(\bm{x_*}) \right),
\label{eq:predictive distribution}
\end{align}
where the predictive mean $\mu(\bm{x}_*)$ and variance $V_f(\bm{x_*})$ can be expressed as
\begin{align}
\label{eq:mean prediction}
\mu(\bm{x}_*)&=\bm{K}_*^T \bm{C}^{-1} \bm{y}, \\
\label{eq:variance prediction}
V_f(\bm{x_*})&={\rm cov}_{tt}^f(\bm{x}_*, \bm{x}_*)-\bm{K} _*^T \bm{C}^{-1}\bm{K}_*.
\end{align}
It can be seen that the mean prediction \Eqref{eq:mean prediction} is a linear combination of the observations $\bm{y}$ , while the variance prediction \Eqref{eq:variance prediction} does not depend on $\bm{y}$. 
The first term in variance, ${\rm cov}_{tt}^f(\bm{x}_*, \bm{x}_*)$, is the prior covariance while the second term is the variance reduction due to the mean prediction. 
For the predictive variance of target observation at $\bm{x}_*$, we can simply add the noise variance $\sigma_t^2$ to that of $f_t(\bm{x}_*)$.

Equation (\ref{eq:mean prediction}) implies that the key feature of multi-output Gaussian process is borrowing strength from a sample of  $q$ source outputs $\{f_1,f_2,...f_q\}$ to predict the target output $f_t$ more precisely. 
This effect is achieved by combining the observed source outputs and target output in a linear form wherein the weight is characterized by covariance matrix $\bm{C}$ and $\bm{K}_{*}$. 
We would like to mention again that the key assumption for the desired function of multi-output Gaussian process is that the source outputs and the target output are correlated, and this correlation can be represented by $\bm{C}$ and $\bm{K}_{*}$.

\subsection{Convolution Process}
\begin{figure}
\centering
\includegraphics[width=1.8in]{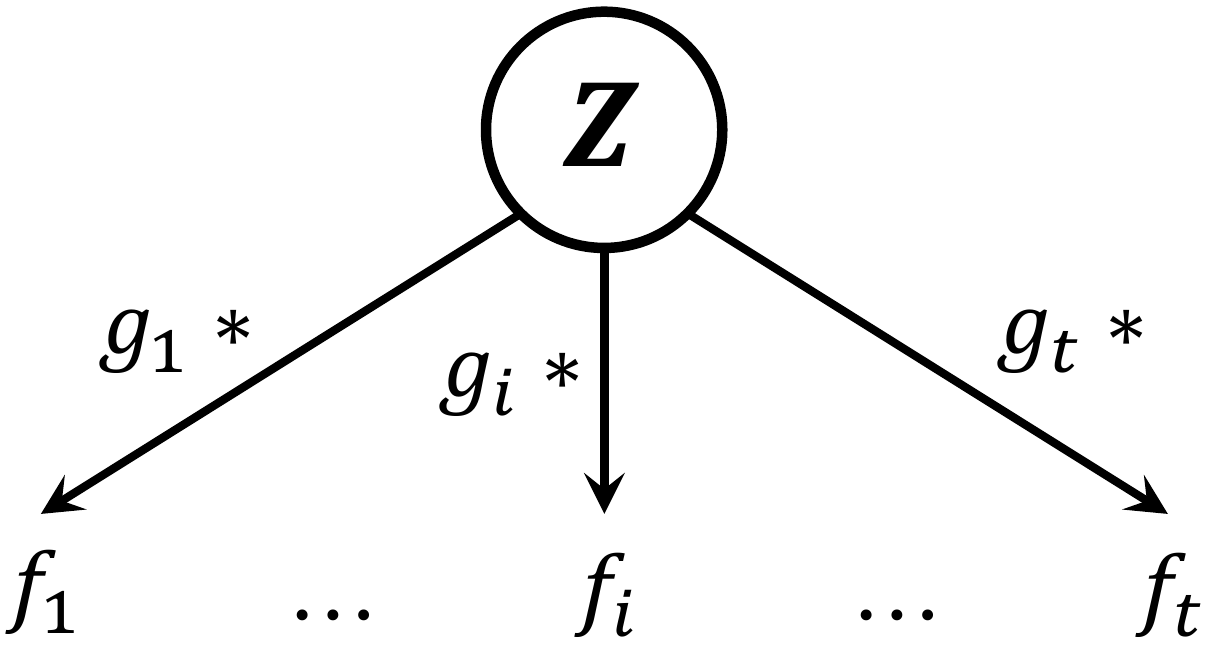}
\caption{Graphical model of convolution process, where $\ast$ denotes a convolution operation.}
\label{fig:convolution process}
\end{figure}

From previous studies \cite{Barry1996, Williams2006}, it is known that the convolution of a Gaussian process and a smoothing kernel is also a Gaussian process. 
Based on this property, we can construct a non-separable generative model which builds valid covariance function for MGP by convolving a base process $Z(\bm{x})$ with a kernel $g(\bm{x})$. 
More precisely, as shown in \Figref{fig:convolution process}, for output $i \in \mathcal{I}$, $f_i(\bm{x})$ can be expressed as
\begin{align}
f_i(\bm{x})=g_i(\bm{x})\ast Z (\bm{x})=\int_{-\infty}^{\infty}g_i (\bm{x}-\bm{u}) Z (\bm{u}) d\bm{u},
\label{eq:single convolution process}
\end{align}
where $\ast$ denotes a convolution operation, $g_i(\bm{x})$ is the output-dependent kernel function and $Z (\bm{x})$ is the shared process across all outputs $f_i(\bm{x}), i \in \mathcal{I}$. 

We assume that  $Z(\bm{x})$ is a commonly used white Gaussian noise process, i.e., ${\rm cov}\left(Z(\bm{x}), Z(\bm{x}^{\prime})\right)=\delta(\bm{x}-\bm{x}^{\prime})$ and $\mathbb{E}(Z(\bm{x}))=0$, where $\delta(\cdot)$ is the Dirac delta function. Note that $f_i(\bm{x})$ is also zero-mean GP, thus the cross covariance can be derived as 
\begin{align}
{\rm cov}_{ij}^f \left(\bm{x}, \bm{x}^{\prime} \right) &={\rm cov}\{ g_i(\bm{x})\ast Z (\bm{x}), g_j(\bm{x}^{\prime})\ast Z (\bm{x}^{\prime})\} \notag\\
&=\int_{-\infty}^{\infty} g_i(\bm{u})g_j(\bm{u}-\bm{v})d \bm{u},
\label{eq:cov in convolution process}
\end{align}
where $\bm{v}=\bm{x}-\bm{x}^{\prime}$. The calculation detail is in Appendix \ref{appendix: covariance derivation} 
Equation (\ref{eq:cov in convolution process}) implies that the correlation between $f_i(\bm{x})$ and $f_j(\bm{x}^{\prime})$ is dependent on the difference  $\bm{x}-\bm{x}^{\prime}$ and the hyperparameters in kernels $g_i$ and $g_j$ when they are constructed by a common process. 

Specially, if we use the Dirac delta function $\delta(\bm{x})$ as the smoothing kernel, i.e., $g_i(\bm{x})=a_i \delta(\bm{x})$, the convolution process will degenerate to the LMC model with single shared latent process, i.e. $f_i(\bm{x})=a_i Z(\bm{x})$ where $a_i \in \mathbb{R}$ is specific to each output $i$ \cite{Alvarez2011}. 
So the convolution process can be considered as a dynamic version of LMC because of the smoothing kernel, which also illustrates the superiority of the non-separable MGP model. 

More generally, we can combine the influence of multiple latent processes and extend \Eqref{eq:single convolution process} to a more flexible version as
\begin{align}
f_i(\bm{x})=\sum_{e=1}^{l}g_{ie}(\bm{x})\ast Z_e (\bm{x}),
\label{eq: multi convolution process}
\end{align}
where $l$ is the number of different latent processes. 
This expression can capture the shared and output-specific information by using a mixture of common and specific latent processes \cite{Fricker2013}. 

\section{Model development}
\label{sec:3-model development}
The proposed framework presents a flexible alternative which can simultaneously reduce negative source information transfer and handle inconsistent input domain. 
In Section \ref{sec:3-model}, our regularized multi-output Gaussian process model is established using a convolution process under the assumption of \emph{consistent} input domain. 
The structure of our model enables the separate information sharing between the target and each source. More importantly, our regularized model can realize the selection of informative sources globally. 
Section \ref{sec:3-statistical} provides some statistical properties for the proposed model, including the consistency and sparsity of estimators.
Section \ref{sec:3-inconsistent} presents the domain adaptation method to deal with the \emph{inconsistent} input domain of sources. 
In Section \ref{sec:3-realization}, we discuss the implementation of our model using Gaussian kernel and $L_1$ norm regularization.

\subsection{Regularized MGCP modeling framework}
\label{sec:3-model}
In this and the following subsection, we focus on the circumstance that the source input domain is consistent with the target input domain. Note that we will relax this assumption in Section \ref{sec:3-inconsistent}.

\begin{figure}
\centering
\includegraphics[width=2.0in]{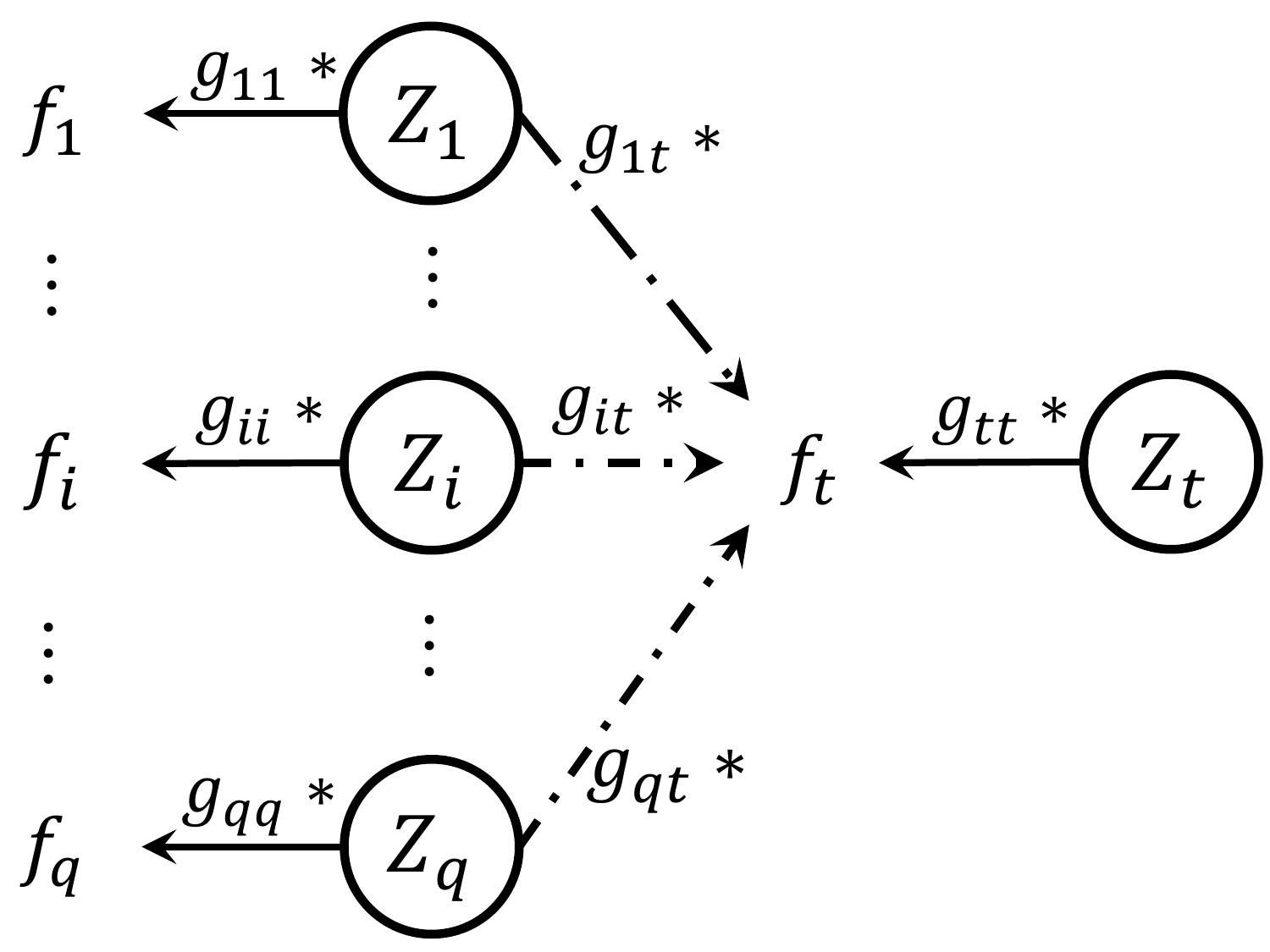}
\caption{The structure of MGP \cite{Kontar2018} for modeling the target output $f_t$.}
\label{fig:structure}
\end{figure}

As described in Section \ref{sec:2-mgp}, we are provided with $q$ source outputs $\{f_i | i\in \mathcal{I}^S\}$, one target output $f_t$, and the observed data $\mathcal{D}=\{\mathcal{D}^S,\mathcal{D}_t\}$. 
Under the framework of MGP, we use CP to construct the covariance functions as shown in \Eqref{eq:cov in convolution process}. The structure of our model is illustrated in \Figref{fig:structure}. 
With the aim of borrowing information from the source outputs to predict the target output more accurately, the latent process $Z_i, i \in \mathcal{I}^S$ and kernels $g_{ii},g_{it}$ serve as the information-sharing channel between the outputs $f_t$ and $f_i$. 
On the other hand, $Z_i$'s are set independent of each other so that no information is shared among source outputs, which significantly reduces the computation complexity that will be analyzed later. 
Considering the existence of target-specific behavior, for simplicity yet without loss of generality, a single latent process $Z_t(\bm{x})$ is added to the construction of $f_t$. 

Based on the structure illustrated in \Figref{fig:structure}, the observation of outputs can be expressed as
\begin{align}
y_i(\bm{x})&=f_i(\bm{x})+\epsilon_i(\bm{x})=g_{ii}(\bm{x})\ast Z_i(\bm{x})+\epsilon_i(\bm{x}), i \in \mathcal{I}^S \notag\\
y_t(\bm{x})&=f_t(\bm{x})+\epsilon_t(\bm{x})=\sum_{j \in \mathcal{I}}g_{jt}(\bm{x}) \ast Z_j(\bm{x})+\epsilon_t(\bm{x}),
\label{eq:structure function}
\end{align}
where $g_{ii}$ is the kernel connecting latent process $Z_i$ and the output $f_i$, and $g_{it}$ is the kernel connecting the latent process $Z_i$ and the target output $f_t$. For the $q$ source outputs, individual kernel for each source enables an accurate approximation for their feature. 
For the target output $f_t$, its shared features with source outputs are encoded in $Z_i$ and $g_{it}$, $i\in \mathcal{I}^S$, while its specific feature is encoded in $Z_t$ and $g_{tt}$.
Based on the assumption that $f(\bm{x})$ is independent with $\epsilon(\bm{x})$, the covariance between any two observations of the outputs $i,j \in \mathcal{I}$ can be decomposed as: 
${\rm cov} \left(y_i(\bm{x}),y_j(\bm{x}^{\prime})\right) 
={\rm cov} \left(f_i(\bm{x}),f_j(\bm{x}^{\prime})\right) + {\rm cov} \left(\epsilon_i(\bm{x}),\epsilon_j(\bm{x}^{\prime}) \right)$. 
To keep the notational consistency, denote ${\rm cov} \left(y_i(\bm{x}),y_j(\bm{x}^{\prime})\right)$ as ${\rm cov}_{ij}^y(\bm{x},\bm{x}^{\prime})$. As $\epsilon_i(\bm{x}), i\in \mathcal{I}$ are i.i.d Gaussian noises, the covariance of two observations $y_i(\bm{x}),y_j(\bm{x}^{\prime})$ can be expressed as
\begin{align}
{\rm cov}_{ij}^y(\bm{x},\bm{x}^{\prime})={\rm cov}_{ij}^f(\bm{x},\bm{x}^{\prime})+\sigma_i^2 \tau_{ij}(\bm{x}-\bm{x}^{\prime}),\ \forall i,j \in \mathcal{I}
\label{eq:covariance decomposition}
\end{align}
where $\tau_{ij}(\bm{x}-\bm{x}^{\prime})$ is equal to $1$ if $i=j$ and $\bm{x}=\bm{x}^{\prime}$, and ${0}$ otherwise. 
Note that every output is a zero-mean GP and $\{Z_i(\bm{x})|,i\in \mathcal{I}\}$ are independent white Gaussian noise processes, so ${\rm cov}_{ij}^f (\bm{x},\bm{x}^{\prime})=0$ for $i,j \in \mathcal{I}^S$ and $i \neq j$, i.e. the covariance across sources are set as zero. And the source-target covariance ${\rm cov}_{it}^f(\bm{x},\bm{x}^{\prime})$ can be calculated as
\begin{align}
{\rm cov}_{it}^f(\bm{x},\bm{x}^{\prime})
=\int_{-\infty}^{\infty} g_{ii}(\bm{u})g_{it}(\bm{u}-\bm{v})d\bm{u},\quad i \in \mathcal{I}^S
\label{eq:cov_it}
\end{align}
where the last equality is based on \Eqref{eq:cov in convolution process}, and $\bm{v}=\bm{x}-\bm{x}^{\prime}$. The detailed calculation can be found in Appendix \ref{appendix: covariance derivation}. In the same way, we can derive the auto-covariance as
\begin{align}
\label{eq:cov_ii}
{\rm cov}_{ii}^f(\bm{x},\bm{x}^{\prime})&=\int_{-\infty}^{\infty} g_{ii}(\bm{u})g_{ii}(\bm{u}-\bm{v})d\bm{u}, i\in \mathcal{I}^S \\
\label{eq:cov_tt}
{\rm cov}_{tt}^f(\bm{x},\bm{x}^{\prime})&=\sum_{j \in \mathcal{I}}\int_{-\infty}^{\infty} g_{jj}(\bm{u})g_{jt}(\bm{u}-\bm{v})d\bm{u}.
\end{align}
Finally based on the above results, we can obtain the explicit expression of covariance matrix $\bm{C}=\bm{K}(\bm{X},\bm{X})+\bm{\Sigma}$ in \Eqref{eq:joint distribution} as
\begin{small}
\begin{align}
\bm{C}=
\renewcommand{\arraystretch}{1.3}
\setlength{\arraycolsep}{1.2pt}
\begin{pmatrix}
\begin{array}{cccc|c}
\bm{C}_{1,1}  & \bm{0} & \cdots & \bm{0} & \bm{C}_{1,t} \\
\bm{0} & \bm{C}_{2,2} & \cdots & \bm{0} & \bm{C}_{2,t} \\
\vdots  & \vdots & \ddots & \vdots & \vdots \\
\bm{0}  & \bm{0} & \cdots & \bm{C}_{q,q} & \bm{C}_{q,t} \\ 
\hline
\bm{C}_{1,t}^T  & \bm{C}_{2,t}^T  & \cdots & \bm{C}_{q,t}^T & \bm{C}_{t,t}
\end{array} 
\end{pmatrix}:=
\begin{pmatrix} \begin{array}{c|c}
\bm{\Omega}_{s,s} & \bm{\Omega}_{s,t} \\ \hline \bm{\Omega}_{s,t}^T & \bm{\Omega}_{t,t} ,
\end{array} \end{pmatrix}
\label{eq:covariance matrix}
\end{align}
\end{small}
where $\bm{C}_{i,j}=\bm{K}_{i,j}+\bm{\Sigma}_{i,j} \in \mathbb{R}^{n_i \times n_j}$ consists of elements $\left\{\bm{C}_{i,j} \right\}_{a,b}={\rm cov}_{ij}^{y}(\bm{x}_{i,a}, \bm{x}_{j,b})$, $1\leq a \leq n_i, 1 \leq b \leq n_j$. 
Re-partition the covariance matrix into four blocks: $\bm{\Omega}_{s,s}$, a block diagonal matrix, representing the covariance of source outputs' data; $\bm{\Omega}_{s,t}$ representing the cross covariance between the source outputs and the target output, which realizes the information transfer from sources to the target; $\bm{\Omega}_{t,t}$ representing the covariance within the target output.

Regarding the structure shown in \Eqref{eq:covariance matrix}, there are two interesting points worth of further discussion. 
The first point is setting the covariance across the source outputs to zero, which is the result of the independency among $\{\bm{Z}_i\}_{i=1}^q$. Ignoring the interactions among sources may cause some  loss of prediction accuracy for the sources, especially when the amount of observed data $n_i$ is small. 
However, we aim to improve the prediction accuracy only for the target and assume the sample data for each source are sufficient, which guarantees the prediction performance of our method. 
Another one is about the covariance between the target and each source, which reveals the advantage of our proposed framework in dealing with negative transfer. 
The source-target covariance function in \Eqref{eq:cov_it} illustrates that $f_t$ can share information with each source  through the kernels  $g_{it}$ and $g_{ii}$ with different hyperparameters.
It can be intuitively understood that if $g_{it}(\bm{x})$ is equal to $0$, the covariance between $f_i$ and $f_t$ is also zero.  
As a result, the prediction of $f_t$ will not be influenced by $f_i$, i.e., no information transfer between them. 
Further, we derive the following theorem which presents the ability of our model in reducing negative transfer.
\newtheorem{theorem}{Theorem}
\begin{theorem}
\label{theorem:conditional distribution}
Suppose that $g_{it}(\bm{x})=0, \forall i \in  \mathcal{U} \subseteq \mathcal{I}^S$ for all $\bm{x}\in \mathcal{X}$. 
For notational convenience, suppose $\mathcal{U}=\{1,2,...,h|h\leq q\}$, then the predictive distribution of the model at any new input $\bm{x}_*$ is unrelated with $\{f_1,f_2,...,f_h\}$ and is reduced to:
\begin{align}
p(y_t(\bm{x}_{*}) | \bm{y})=\mathcal{N}(&\bm{k}_{+}^T \bm{C}_{+}^{-1} \bm{y}_{+}, \notag\\
&{\rm cov}_{tt}^f(\bm{x}_{*},\bm{x}_{*})+\sigma_t^2-\bm{k}_{+}^T \bm{C}_{+}^{-1} \bm{k}_{+}), \nonumber
\end{align}
where $\bm{k}_{+}=(\bm{K}_{h+1,*}^T,...,\bm{K}_{q,*}^T,\bm{K}_{t,*}^T)^T$, $\bm{y}_{+}=(\bm{y}_{h+1}^T,...,\bm{y}_{q}^T,\bm{y}_{t}^T)^T$, and 
\begin{small}
$$\bm{C}_{+}=
\renewcommand{\arraystretch}{1.2}
\setlength{\arraycolsep}{1.2pt}
\begin{pmatrix}
\bm{C}_{h+1,h+1} & \cdots & \bm{0} & \bm{C}_{h+1,t} \\
\vdots & \ddots & \vdots & \vdots  \\
\bm{0}  & \cdots & \bm{C}_{q,q}  & \bm{C}_{q,t} \\
\bm{C}_{h+1,t}^T & \cdots & \bm{C}_{q,t}^T & \bm{C}_{t,t}
\end{pmatrix}.$$
\end{small}
\end{theorem}
The proof is detailed in Appendix \ref{appendix: conditional distribution}. 
This theorem demonstrates one key property of our framework. If we penalize the smoothing kernels $\{g_{it}(\bm{x})\}_{i \in  \mathcal{U}}$ to zero, the MGCP is actually reduced to a marginalized version, which only contains source outputs $\{g_{it}(\bm{x})\}_{i \in \mathcal{I}^S / \mathcal{U}}$ and the target output. 
The possible negative transfer between $\{f_{i}\}_{i \in  \mathcal{U}}$ and $f_t$ can thus be avoided completely. This result is based on the fact that if $g_{it}(\bm{x})=0$, 
$${\rm cov}_{it}^f (\bm{x})=\int_{-\infty}^{\infty} g_{ii}(\bm{u})g_{it}(\bm{u}-\bm{v})d\bm{u}=0,$$
and $\bm{C}_{it}=0$. 

To apply the idea in \autoref{theorem:conditional distribution} to model regularization, we denote that $g_{it}(\bm{x})=\theta_{i0} \tilde{g}_{it}(\bm{x})$, where $\theta_{i0}$ satisfies the condition that $g_{it}(\bm{x})=0, \forall \bm{x}$ if and only if $\theta_{i0}=0$. 
Let $\bm{\theta}$ be the collection of all parameters in the model and $\bm{\theta}_0=\{\theta_{i0}| i \in \mathcal{I}^S\} \subset \bm{\theta}$ . 
Then, based on the results of \autoref{theorem:conditional distribution}, our regularized model can be derived as:
\begin{align}
\underset{\bm{\theta}}{\rm max}\ L_{\mathbb{P}}(\bm{\theta}| \bm{y})
=&L(\bm{\theta}| \bm{y})-\mathbb{P}_{\gamma}(\bm{\theta}_0)\notag\\
=&-\frac{1}{2}\bm{y}^T\bm{C}^{-1} \bm{y}-\frac{1}{2}{\rm log}|\bm{C}| \notag\\
&-\frac{N}{2}log(2\pi)-\mathbb{P}_{\gamma}(\bm{\theta}_0),
\label{eq:regularized log-likelihood}
\end{align}
where $L_{\mathbb{P}}(\bm{\theta}| \bm{y})$ denotes the regularized log-likelihood, $L(\bm{\theta}| \bm{y})$ denotes the normal log-likelihood for Gaussian distribution, and $\mathbb{P}_{\gamma}(\bm{\theta}_0)$ is a non-negative penalty function. 
To make the smoothing kernel connecting target and uncorrelated source to $0$, common choices of the regularization function include: $L_1$ norm $\mathbb{P}_{\gamma}(\bm{\theta}_0)=\gamma |\bm{\theta}_0|$ and smoothly clipped absolute deviation (SCAD) function \cite{Fan2001}. 
The validity of our method is ensured by two claims. 
Firstly, based on the theory of multivariate Gaussian distribution, if the source $f_i$ is uncorrelated with the target $f_t$, then the corresponding covariance matrix block $\bm{C}_{it}$ should be zero. 
Secondly, \autoref{theorem:conditional distribution} guarantees that by shrinking some elements of $\bm{\theta}_0$, $\{\theta_{i0}\}_{i\in \mathcal{U}}$, to zero, $\{\bm{C}_{it}\}_{i\in \mathcal{U}}=\bm{0}$ and the target output can be predicted without the influence of the source outputs $\{f_i\}_{i \in \mathcal{U}}$. 
Another unique advantage of the proposed method is that it is a global regularized model, since the shrinkage of parameters are applied to all the sources simultaneously, which is different from the local regularization over a subset of data in \cite{Kontar2020}.

Besides the property of global regularization over all the sources, the computational complexity of our method in parameter optimization  is greatly reduced because of the sparse covariance matrix. 
Based on the partitioned covariance matrix
$\bm{C}=
\begin{pmatrix} \bm{\Omega}_{s,s} & \bm{\Omega}_{s,t} \\ \bm{\Omega}_{s,t}^T & \bm{\Omega}_{t,t} \end{pmatrix}$ and using the inversion lemma of a partitioned matrix, the log-likelihood function can be decomposed as:
\begin{align}
L(\bm{\theta}| \bm{y})=&-\frac{1}{2}\left[\bm{\tilde{y}}^T \bm{\Omega}_{s,s}^{-1}\bm{\tilde{y}}+(\bm{A} \bm{\tilde{y}}-\bm{y}_t)^T \bm{B}^{-1}(\bm{A}\bm{\tilde{y}}-\bm{y}_t) \right] \notag\\
&-\frac{1}{2} \left[ \log |\bm{\Omega}_{s,s}|+ \log |\bm{B}| \right]-\frac{N}{2}log(2\pi),
\end{align}
where $\bm{\tilde{y}}=\{\bm{y}_1^T,...,\bm{y}_q^T\}^T$, $\bm{A}=\bm{\Omega}_{s,t}^T \bm{\Omega}_{s,s}^{-1}$, $\bm{B}= \bm{\Omega}_{t,t}-\bm{A}\bm{\Omega}_{s,t}$ is the Schur complement. 
The computational load of MLE is mainly on calculating the inverse of covariance matrix $\bm{\Omega}_{s,s}$ and $\bm{B}$. As $\bm{\Omega}_{s,s}$ is a diagonal blocked matrix with $q$ square matrix $\bm{C}_{i,i} \in \mathbb{R}^{n \times n}$, the complexity for $\bm{\Omega}_{s,s}^{-1}$ is $O(qn^3)$. 
As $\bm{B}\in \mathbb{R}^{n_t\times n_t}$, the complexity for $\bm{B}^{-1}$ is $O(n_t^3)$. 
As a result, the complexity of our method is $O(qn^3+n_t^3)$. 
\textcolor{black}{
However, in the ordinary MGP methods \cite{Alvarez2011}, $\bm{C}$ is a full matrix without zero blocks, so the complexity increases to $O((qn)^3)$. Therefore, the whole computational complexity is $O((qn)^3+n_t^3)$. The above complexity calculation still holds when some sources have different input domains with the target, which will be analyzed in Section \ref{sec:3-inconsistent}.}

\subsection{Statistical properties for regularized MGCP}
\label{sec:3-statistical}
In Section \ref{sec:3-model}, we have discussed that if the source output $f_i$ is uncorrelated with the target output $f_t$, the cross-covariance between them should be zero, i.e. $\bm{C}_{i,t}=\bm{0}$.
On the other hand, if the kernel $g_{it}(\bm{x})=0$, then $\bm{C}_{i,t}=\bm{0}$ and thus the predictive distribution of $f_t(\bm{x}_*)$ is uncorrelated with the observations from the source output $f_i$ according to \autoref{theorem:conditional distribution}. 
Therefore, to avoid negative transfer, the estimated parameter $\hat{\theta}_{i0}$ should be zero, which can be realized through the regularized estimation.

In this subsection, we provide some asymptotic properties of the regularized maximum likelihood estimator $\hat{\bm{\theta}}$. 
Same as the  last subsection, suppose there are $q$ elements in $\bm{\theta}_0$, denoted by $\{\theta_{10},\theta_{20},...,\theta_{q0}\}$, which correspond to the $q$ smoothing kernels $\{g_{1t},g_{2t},...,g_{qt}\}$ respectively. 
Denote the true parameter values of $\bm{\theta}_0$ and $\bm{\theta}$ in \Eqref{eq:regularized log-likelihood} as $\bm{\theta}_0^*$ and $\bm{\theta}^*$. Suppose there are $h$ zero elements in $\bm{\theta}_0^*$. 
 Regarding to the penalty function, we assume that $\mathbb{P}_{\gamma}(\bm{\theta}_0) \geq 0, \forall \bm{\theta}_0$; $\mathbb{P}_{\gamma}(\bm{0}) = 0$;  $\mathbb{P}_{\lambda}(\bm{\theta}_0^{\prime}) \geq \mathbb{P}_{\lambda}(\bm{\theta}_0) $ if $ |\bm{\theta}_0^{\prime}| \geq |\bm{\theta}_0|$. 
 These typical assumptions are easily satisfied by the previously mentioned penalty functions.

Before discussing the statistical properties of the regularized model \Eqref{eq:regularized log-likelihood}, we first need to introduce the consistency of the maximum log-likelihood estimator (MLE), $\hat{\bm{\theta}}_{\#}$, for the unpenalized $L(\bm{\theta}|\bm{y})$. 
Note that the observations of Gaussian process are dependent. 
So based on some regularity conditions for stochastic process, it has been proved that $\hat{\bm{\theta}}_{\#}$ asymptotically converges to $\bm{\theta}^*$ with rate $r_N$ s.t. $r_N \rightarrow \infty$ as $N\rightarrow \infty$, i.e.,
\begin{align}
\| \hat{\bm{\theta}}_{\#}-\bm{\theta}^*\| = O_P(r_N^{-1}).
\label{non-penalized maximum log-likelihood estimator consistency}
\end{align} 
For more details of the regular conditions and consistency proof, please refer to Appendix \ref{appendix:regularity conditions} and the chapter 7 in \cite{Basawa2014}. 

We first discuss the consistency of the MLE for the regularized log-likelihood $L_{\mathbb{P}}(\bm{\theta}| \bm{y})$. 
\begin{theorem}
\label{theorem:consistency}
Suppose that the MLE for $L(\bm{\theta}|\bm{y})$, $\hat{\bm{\theta}}_{\#}$, is $r_N$ consistent, i.e., satisfying \Eqref{non-penalized maximum log-likelihood estimator consistency}. If $\max \{ |\mathbb{P}^{\prime \prime}_{\gamma}({\theta}_{i0}^*)|: {\theta}_{i0}^* \neq 0\} \rightarrow 0$, then there exists a local maximizer $\hat{\bm{\theta}}$ of  $L_{\mathbb{P}}(\bm{\theta}|\bm{y})$ s.t. $\|\hat{\bm{\theta}}-\bm{\theta}^{*} \|=O_P(r_N^{-1}+r_0)$, where $r_0=\max \{ |\mathbb{P}^{\prime}_{\gamma}({\theta}_{i0}^*)|: {\theta}_{i0}^* \neq 0\}$.
\end{theorem}

The proof is detailed in Appendix \ref{appendix:consistency theorem}. 
This theorem states that if the derivative of penalty function satisfies some conditions, the estimator of the regularized log-likelihood is also consistent. 
If we take a proper sequence of $\gamma$ for the penalty, for example choose $\gamma$ to make $r_0$ satisfies $r_0=o_P(r_N^{-1})$, $\hat{\bm{\theta}}$ is also $r_N$ consistent as $\hat{\bm{\theta}}_{\#}$. 
The condition in this theorem, $\max \{ |\mathbb{P}^{\prime \prime}_{\gamma}(\bm{\theta}_{i0})|: \bm{\theta}_{i0} \neq 0\} \rightarrow 0$, is easily satisfied for common regularization functions. 
For example, if $\mathbb{P}_{\gamma}(\theta_{i0})=\gamma|\theta_{i0}|$, then $|\mathbb{P}^{\prime \prime}_{\gamma}(\bm{\theta}_{i0})|=0$ satisfies. 

Besides the consistency, another key property of $\hat{\bm{\theta}}$ is that it possess the sparsity, which is provided in \autoref{theorem:sparsity} as follows.
\begin{theorem}
\label{theorem:sparsity}
Let $\bm{\theta}_{10}^{*}$ and $\bm{\theta}_{20}^{*}$ contain the zero and non-zero components in  $\bm{\theta}_{0}^{*}$ respectively.
Assume the conditions in \autoref{theorem:consistency} also hold, and $\hat{\bm{\theta}}$ is $r_N$ consistent by choosing proper $\gamma$ in $\mathbb{P}_{\gamma}(\bm{\theta}_0)$. If 
$\underset{N \rightarrow \infty}{\lim \inf}\ \underset{\theta \rightarrow 0^+}{\lim \inf}\ \gamma^{-1}\mathbb{P}^{\prime}_{\gamma}(\theta) >0$ and $(r_N \gamma)^{-1} \rightarrow 0$, 
then $$\underset{N \rightarrow \infty}{\lim}P \left( \hat{\bm{\theta}}_{10}=\bm{0} \right)=1.$$
\end{theorem}
The proof is detailed in Appendix \ref{appendix:sparsity theorem}. 
This theorem implies that by choosing proper penalty functions and tuning parameters, the regularized MGCP model can realize variable selection, i.e. the estimator $\hat{\bm{\theta}}$ can perform as well as if ${\bm{\theta}}_{10}=\bm{0}$ is known in advance. More importantly, in our model, the variable selection of $\bm{\theta}$ means the  selection of informative source based on \autoref{theorem:conditional distribution}. 
The conditions in this theorem can also be satisfied easily. Again taking the example $\mathbb{P}_{\gamma}(\theta_{i0})=\gamma|\theta_{i0}|$, if we let $\gamma=r_N^{-1/2}$, then $\underset{N \rightarrow \infty}{\lim \inf}\ \underset{\theta \rightarrow 0^+}{\lim \inf}\ \gamma^{-1}\mathbb{P}^{\prime}_{\gamma}(\theta)=1$ and $(r_N\gamma)^{-1}=r_N^{-1/2}\rightarrow 0$, which satisfies the conditions. 

\subsection{Domain adaptation through marginalization and expansion (DAME)}
\label{sec:3-inconsistent}
The discussions in Section \ref{sec:3-model} and \ref{sec:3-statistical} are based on the assumption that the target and source data share the same input domain. 
However, as we emphasized in the introduction, domain inconsistency is a common issue in transfer learning. 
In this subsection, we propose an effective domain adaptation method for dealing with the domain inconsistency in our MGCP model. 
The general assumption for the proposed domain adaptation method is that there is at least one commonly shared input feature between each  source and the target. 
But we do not require that all sources share the same input, i.e., different sources can share different dimensions with the target.

\begin{figure}[!t]
\centering
\subfloat[]{\includegraphics[width=1.5in]{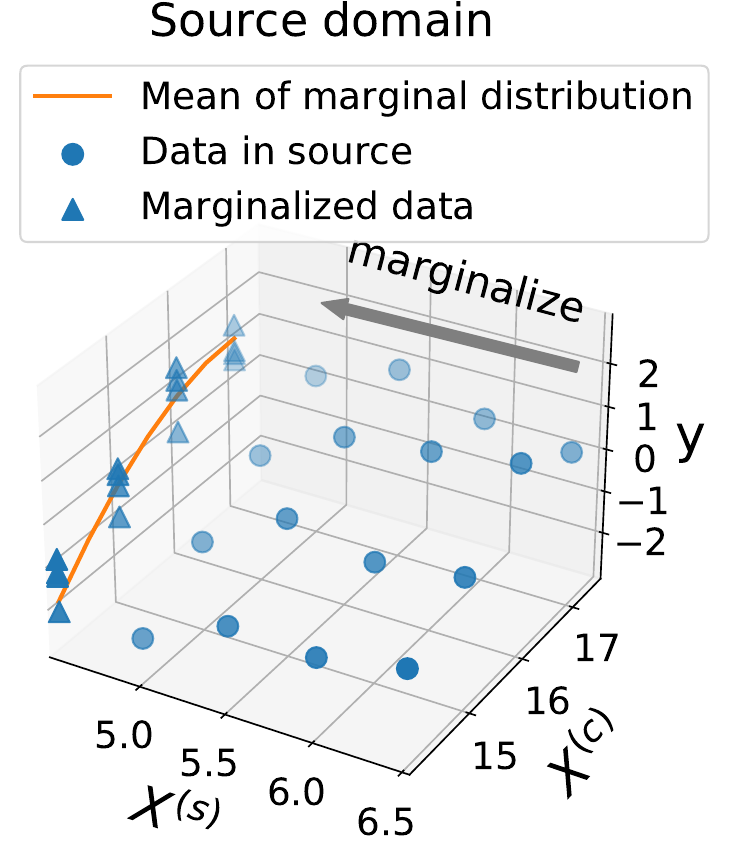}
\label{fig:domain adaptation 1}}
\hfil
\subfloat[]{\includegraphics[width=1.5in]{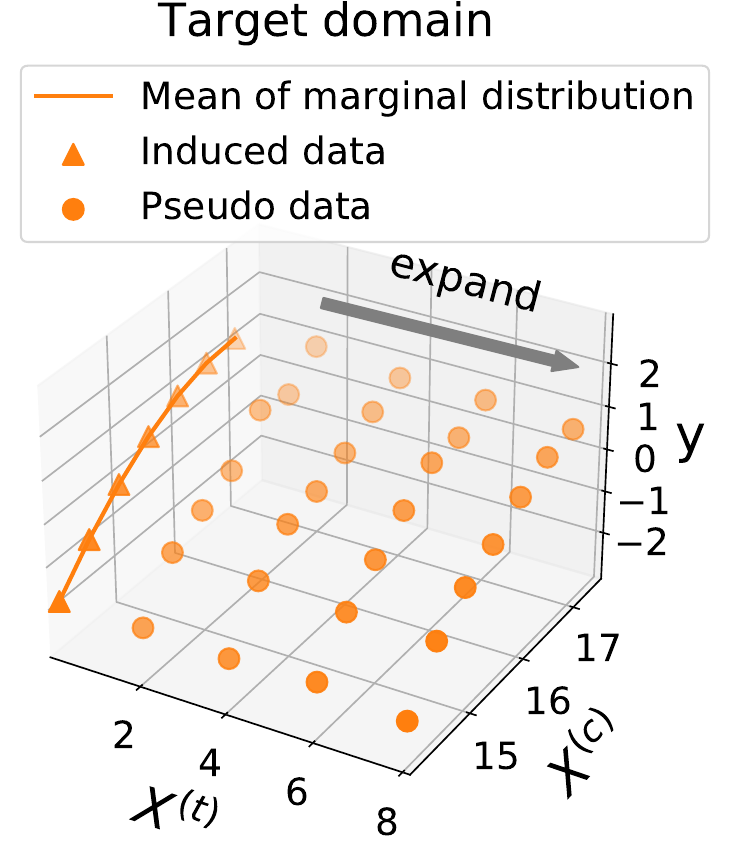}
\label{fig:domain adaptation 2}}
\caption{Illustration of the marginalization based domain adaptation using the normalized density data of ceramic product. (a). Marginalize source data to the domain only with feature $x^{(c)}$, and obtain marginal distribution through kernel regression; (b). Induce data based on the marginal distribution and expand them to the target domain.}
\label{fig:domain adaptation}
\end{figure}

The basic idea of our domain adaptation method is to first marginalize extra features in the sources, then expand missing features to align with the target input domain. 
More specifically, our method aims to find the marginal distribution of the source data in the shared input domain  with the target,  then create a pseudo dataset in the target input domain. Thus, we name the method as DAME.
This newly created pseudo dataset will be in the same input domain as the target data and have the same marginal distribution as the original source data, which can be used as the new source data to plug in the proposed MGCP model. 
Figure \ref{fig:domain adaptation} shows the adaptation procedure using the normalized density data of ceramic product, where the source input domain contains two features $\bm{x}^{(c)}$ and $\bm{x}^{(s)}$, and the target input domain contains features $\bm{x}^{(c)}$ and $\bm{x}^{(t)}$. 
In  \Figref{fig:domain adaptation 1}, the source data are marginalized to the domain which only has feature $\bm{x}^{(c)}$, and a marginal distribution is obtained based on the marginalized data. 
In  \Figref{fig:domain adaptation 2}, several data are induced according to the marginal distribution, then we expand them to get the pseudo data which have the same features with the target data. 

To generalize the example in \Figref{fig:domain adaptation}, we slightly abuse the notation and focus on one source $\mathcal{D}_i=\{\bm{X}_i,\bm{y}_i\}, i\in \mathcal{I}^S$. 
Note that the proposed method will be applied to every source that does not have consistent domain with the target. Let $\bm{x}^{(c)} \in \mathbb{R}^{d_c}$ denote the shared features in both the target and source input domain, $\bm{x}^{(s)} \in \mathbb{R}^{d_i}$ denote the unique features in the input domain of the $i$th source, and $\bm{x}^{(t)} \in \mathbb{R}^{d_t}$ denote the unique features in the target input domain. 
Then, any source and target data can be expressed as
$$\bm{x}_{i,\cdot}=\begin{pmatrix}  \bm{x}_{i,\cdot}^{(c)} \\ \bm{x}_{i,\cdot}^{(s)} \end{pmatrix}, 
\bm{x}_{t,\cdot}=\begin{pmatrix}  \bm{x}_{t,\cdot}^{(c)} \\ \bm{x}_{t,\cdot}^{(t)} \end{pmatrix}.$$

The first step is to marginalize the extra features. Define the shared input domain as $\mathcal{X}^P$ which is represented by $\bm{x}^{(c)}$, and a projection matrix 
$\bm{P}=\begin{pmatrix} \bm{I}_{d_c} & \bm{0}_{d_c \times d_i} \end{pmatrix}$. 
Then, we can get marginalized source data $\mathcal{D}_i^{P}=\{ \bm{X}_i^P, \bm{y}_i \}$, where $\bm{X}_i^P=\bm{P}\bm{X}_i =\left(\bm{x}^{(c)}_{i,1},...,\bm{x}^{(c)}_{i,n_i}\right)$. 
The projected data usually get too dispersed in the shared domain, e.g., the blue triangle in \Figref{fig:domain adaptation}, and the dispersion will be recognized as large measurement noise of the data. 
As a result, a smoothing method is needed to extract the overall trend of the marginalized data and generate induced data with smaller dispersion. 
Many non-parametric methods are available for this purpose, such as kernel regression, B-spline, and GP model, etc. 
In our work, kernel regression is chosen to model the marginalized data, and $n_{i^{\prime}}$ samples $\{\bm{x}_{i^{\prime}, a}^{(c)}, {y}_{i^{\prime}, a}\}_{a=1}^{n_{i^{\prime}}}$ are induced based on the trained model
\begin{align}
{y}_{i^{\prime}, a}=\frac{\sum_{b=1}^{n_i}K_{\lambda}(\bm{x}^{(c)}_{i,b},\bm{x}_{i^{\prime}, a}^{(c)})y_{i,b}}{\sum_{b=1}^{n_i}K_{\lambda}(\bm{x}^{(c)}_{i,b},\bm{x}_{i^{\prime}, a}^{(c)})},
\label{eq:induce data}
\end{align}
where $K_{\lambda}$ is the kernel function and $\lambda$ is the estimated hyperparameter through cross-validation. 
Note we use $i^{\prime}$ to denote a new (marginalized) source resulting from the original source $i$. 
For example, the mean of marginal distribution is represented by the orange curve in \Figref{fig:domain adaptation}, and the induced data are represented by the orange triangle in \Figref{fig:domain adaptation 2}.

The second step of DAME is to expand the $\{\bm{x}_{i^{\prime}, a}^{(c)}, {y}_{i^{\prime}, a}\}_{a=1}^{n_{i^{\prime}}}$ to include the unique features in the target domain, i.e. $\bm{x}^{(t)}$. To realize this idea, we expand the marginalized data along the dimension of $\bm{x}^{(t)}$ by adding i.i.d noise which simulates the measurement error. 
For example, if $\bm{x}^{(t)}$ is one-dimensional and the observed target data have lower bound $\bm{x}_{\rm low}^{(t)}$ (e.g., $\bm{x}_{\rm low}^{(t)}$=2 in \Figref{fig:domain adaptation 2}) and upper bound $\bm{x}_{\rm up}^{(t)}$ (e.g., $\bm{x}_{\rm low}^{(t)}$=8 in \Figref{fig:domain adaptation 2}), choose $n_{i^{\prime \prime}}$ values, $\{\bm{x}_ {i^{\prime \prime},b}^{(t)}\}_{b=1}^{n_{i^{\prime\prime}}}$, spaced in $\left[\bm{x}_{\rm low}^{(t)}, \bm{x}_{\rm up}^{(t)}\right]$. 
The pseudo data of the source $i$ can be expressed as:
\begin{align}
\mathcal{D}_{i}^{\rm new}=\left\{ \begin{pmatrix} \bm{x}_{i^{\prime}, a}^{(c)} \\ \bm{x}_{i^{\prime \prime}, b}^{(t)} \end{pmatrix}, {y}_{i^{\prime\prime}, a, b} \mid a\in [1, n_{i^{\prime}}], b \in [1, n_{i^{\prime \prime}}] \right\},
\label{eq:pseudo data}
\end{align}
where ${y}_{i^{\prime\prime},a, b}={y}_{i^{\prime},a}+\epsilon_{a,b}$ and $\epsilon_{a,b}$ is an i.i.d Gaussian noise. For the standard deviation of $\epsilon_{a,b}$, we can choose the estimated noise deviation of target data using a single-output GP.
In \Figref{fig:domain adaptation 2}, we take $n_{i^{\prime \prime}}=4$ and the pseudo data is represented by the orange dots.
This expanding approach is intuitively practicable as no prior information about the missing features for source data is given. 
If domain knowledge is available for the expansion step, e.g., there is a monotonically increasing trend along $\bm{x}^{(t)}$ in \Figref{fig:domain adaptation 1}, it is also straightforward to incorporate such information.

For the pseudo dataset, our DAME method preserves the marginal information of sources in the shared domain and does not introduce other information in the target-specific domain. This is the unique property of our DAME method. This method will benefit the target if the pseudo dataset is informative. On the other hand, even if the pseudo dataset provides negative information on the target prediction after the marginalization and expansion, our regularized model can identify it and exclude in the learning process. It is worth noting that the existing domain adaptation methods by minimizing the difference between the transformed source and target data might not be able to mitigate the potential negative transfer. Besides, finding an optimal feature mapping (like the existing methods) for the source and target within MGCP training would be extremely difficult, if not impossible. 

\textcolor{black}{
In addition, it is worth analyzing the complexity of our method under the circumstance of inconsistent input domains. For the domain adaptation process, the main computational load is on applying the smooth method to obtain the marginal distribution. It will not exceed $O(n^3)$ for each source whether we use kernel regression or GP model. Thus, the complexity of domain adaptation process is not larger than that of constructing the MGCP model in our method. Therefore, the computational complexity of our method is still $O(qn^3+n_t^3)$ when domain inconsistency occurs.
}

\subsection{Implemetation using Gaussian kernel and $L_1$ norm}
\label{sec:3-realization}
In this subsection, we use Gaussian kernel to implement the modeling framework introduced in Section \ref{sec:3-model}-\ref{sec:3-inconsistent}. 
Gaussian kernel is a very popular choice which is flexible for various spatial characteristics with a small number of hyperparameters. In order to obtain a neat closed form of the convolved covariance function, we take the smoothing kernel as:
\begin{align}
g_{ij}(\bm{x})=\alpha_{ij} \pi^{-\frac{d}{4}}|\bm{\Lambda}|^{-\frac{1}{4}}\exp \left(-\frac{1}{2}\bm{x}^T \bm{\Lambda}^{-1}\bm{x} \right)\  i,j \in \mathcal{I}
\label{eq:smooth kernel},
\end{align}
where $\alpha_{ij}$ is the scaling parameter and $\bm{\Lambda}$ is the diagonal matrix representing the length-scale for each input feature.
Using the domain adaptation introduced in Section \ref{sec:3-inconsistent}, we can assume that every source $i \in \mathcal{I}^S$ is transformed to have the same input domain as the target.
By plugging the kernel \Eqref{eq:smooth kernel} in \Eqref{eq:cov_it}-\Eqref{eq:cov_tt} we obtain
\begin{small}
\begin{align}
&{\rm cov}_{it}^f(\bm{x},\bm{x}^{\prime})
= 2^{\frac{d}{2}}\alpha_{ii}\alpha_{it}\frac{|\bm{\Lambda}_{ii}|^{\frac{1}{4}} |\bm{\Lambda}_{it}|^{\frac{1}{4}}} {|\bm{\Lambda}_{ii}+\bm{\Lambda}_{it}|^{\frac{1}{2}}} \times \notag\\
&\qquad \qquad \qquad \exp \left[-\frac{1}{2}(\bm{x}-\bm{x}^{\prime})^T (\bm{\Lambda}_{ii}+\bm{\Lambda}_{it})^{-1}(\bm{x}-\bm{x}^{\prime}) \right], \notag\\
&{\rm cov}_{ii}^f(\bm{x},\bm{x}^{\prime})
= \alpha_{ii}^2 \exp \left[-\frac{1}{4}(\bm{x}-\bm{x}^{\prime})^T \bm{\Lambda}_{ii}^{-1}(\bm{x}-\bm{x}^{\prime}) \right], \notag\\
&{\rm cov}_{tt}^f(\bm{x},\bm{x}^{\prime})
=\sum_{j \in \mathcal{I}} \alpha_{jt}^2 \exp \left[-\frac{1}{4}(\bm{x}-\bm{x}^{\prime})^T \bm{\Lambda}_{jt}^{-1}(\bm{x}-\bm{x}^{\prime})\right], 
\label{eq:cov collection}
\end{align}
\end{small}
where $i \in \mathcal{I}^S$. \Eqref{eq:cov collection} shows that by using the kernel in \Eqref{eq:smooth kernel}, the covariance functions are similar to the traditional Gaussian kernel, especially for the auto-covariance within each source.

Then, based on these results,  in regularized log-likelihood of \Eqref{eq:regularized log-likelihood}, the collection of all parameters is $\bm{\theta}=\left\{ \alpha_{ii}, \alpha_{it}, \bm{\Lambda}_{ii}, \bm{\Lambda}_{it}, \sigma_i | i\in \mathcal{I} \right\}$, and the sparsity parameters is $\bm{\theta}_0=\{\alpha_{it}| i\in \mathcal{I}^S\}$. 
Moreover, if we take $L_1$ norm as the regularization function and consider the transformed source data, \Eqref{eq:regularized log-likelihood} will become
\begin{small}
\begin{align}
\underset{\bm{\theta}}{\rm max}\ L_{\mathbb{P}}(\bm{\theta}| \bm{y}^{\rm new})=&L(\bm{\theta}| \bm{y}^{\rm new})-\gamma \sum_{i=1}^{q}|\alpha_{it}|,
\label{eq:regularized log-likelihood with L1}
\end{align}
\end{small}
where $\bm{y}^{\rm new}=\{\bm{y}^{\rm nda},\bm{y}^{\rm da}\}$, $\bm{y}^{\rm nda}$ includes the data from sources which do not conduct domain adaptation, $\bm{y}^{\rm da}$ includes the data from sources conducting domain adaptation; $\gamma$ is the tuning parameter and has critical effect on the optima. 
Typically, the choice of tuning parameter is made through a grid search with cross validation(CV) or generalized CV, such as leave-one-out  (generalized) CV and 5-fold (generalized) CV.

The estimated parameter $\bm{\hat{\theta}}$ are obtained through solving the problem in \Eqref{eq:regularized log-likelihood with L1}.  
However, there are two noticeable details in practice. 
Firstly, as this optimization problem is not a convex problem and multiple local optima exist with high probability, we usually need to set random initial values for several times. 
Secondly, as the commonly used gradient methods, such as L-BFGS method and conjugate gradient method, require the objective function to be smooth, they cannot be applied to solving \Eqref{eq:regularized log-likelihood with L1} because $L_1$ norm function is not smooth at zero point. 
To solve this issue, we take a Huber smooth approximation as
\begin{small}
\begin{align}
\label{eq:smooth approximation}
\gamma \sum_{i=1}^{q}|\alpha_{it}| \approx
\gamma \sum_{i=1}^{q}
\left\{
\begin{aligned}  
\frac{1}{2\eta}\alpha_{it}^2, |\alpha_{it}|\leq \eta \\
|\alpha_{it}|-\frac{\eta}{2}, |\alpha_{it}|> \eta
\end{aligned}
\right.
\end{align}
\end{small}
where $\eta$ is a small constant, e.g. $10^{-4}$. 
As the maximum bias between the approximation and original function is $\eta$, it brings little influence to the optima and makes the common gradient method applicable. 
Finally for prediction, calculate $\bm{C}$, $\bm{K}_*$ and ${\rm cov}_{tt}^f(\bm{x}_*,\bm{x}_*)$ in \Eqref{eq:mean prediction}-\Eqref{eq:variance prediction} with $\bm{\hat{\theta}}$ at point $\bm{x}_*$. 
Then, the predictive distribution of $f_t(\bm{x}_*)$ is in the form of \Eqref{eq:predictive distribution}.

The implementation of the regularized MGCP modeling in this work is summarized in Algorithm \autoref{algorithm: summary}.
\begin{algorithm}
\caption{Regularized MGCP model with marginalizing-expanding domain adaptation}
\label{algorithm: summary}
\begin{algorithmic}[1]
\Require Sources data $\{\mathcal{D}_i\}_{i=1}^q$, target data $\mathcal{D}_t$, $\gamma$, $\eta$, $\bm{x}_*$
\For{source $\mathcal{D}_i$ with inconsistent input domain}
	\State Obtain marginalized data $\{\bm{x}_{i,a}^{(c)}, \bm{y}_{i,a}\}_{i=1}^{n_i}$, where $\bm{x}^{(c)}$
	\Statex \quad \enspace are the shared features.
	\State Obtain the mean of marginal distribution through 
	\Statex \quad \enspace training a kernel regression model. Induce data in 
	\Statex \quad \enspace the shared domain using \Eqref{eq:induce data}. 
	\State Generate pseudo dataset $\mathcal{D}_{i}^{\rm new}$ by expanding the 
	\Statex \quad \enspace induced data to the target domain  through \Eqref{eq:pseudo data}. 
\EndFor
\State Generate random start point $\bm{\theta}_{start}$.
\State Obtain estimator $\hat{\bm{\theta}}$ through solving the optimization problem \Eqref{eq:smooth approximation}, where the covariance matrix is calculated based on \Eqref{eq:cov collection}. 
\State Calculate $f_t(\bm{x}_*)$ using \Eqref{eq:predictive distribution}-(\ref{eq:variance prediction}) with $\hat{\bm{\theta}}$.
\Ensure $\hat{\bm{\theta}}$, $f_t(\bm{x}_*)$
\end{algorithmic}
\end{algorithm}

\color{black}
\subsection{Unique Methodology Contribution}
As mentioned in the introduction, the works of  \cite{Kontar2020},\cite{Kontar2018} also focus on predicting one output through multi-output Gaussian process. Although the covariance structure in \cite{Kontar2018} is similar to us, the work in \cite{Kontar2018} mainly focuses on realizing the transfer from multiple sources to one target. Moreover, works in \cite{Kontar2018} does not consider negative transfer and the corresponding theoretical guarantee on the regularization, which is the major focus in our work.

For the two-stage strategy in \cite{Kontar2020}, it conducts regularization pair-wisely between the target and each source, and combines each sub-model's prediction linearly with different weights. This way has two main drawbacks. 
First, the correlation in one pair might be influenced by other sources. In other words, the strong correlation in one pair might be the results of other sources, and such strong correlation might disappear when considering all sources together. 
Second, the integration of all pairs is conducted by the predictive variance, which is a sub-optimal way for both performance and interpretability.

Finally, compared with existing MGP models, this work is the first one considering inconsistent input domain problem. This technique can increase available source data for transfer learning and multi-task learning with MGP. 

\color{black}
\section{Numerical studies}
\label{sec:4-numerical studies}
We apply the proposed regularized multi-output Gaussian convolution process model, referred as MGCP-R, to two simulation cases and one real case. In Section \ref{sec:4-general settings}, we introduce the general settings and benchmark methods for our numerical studies. Section \ref{sec:4-simulation 1} demonstrates the advantages of our method in reducing negative transfer when the sources have the consistent input domain with the target. Section \ref{sec:4-simulation 2} presents the effectiveness of our framework in dealing with the inconsistent source input domain. And in \ref{sec:4-simulation 3}, we test and verify the performance with a moderate number of sources and input dimensions. Finally in Section \ref{sec:4-real case}, we apply the proposed modeling framework to the density prediction of ceramic product.

\subsection{General settings}
\label{sec:4-general settings}
In this section, we discuss the general settings for assessing the benefits of MGCP-R using simulated data. To evaluate the performance in selecting informative sources and mitigating the negative transfer of knowledge, we randomly generate observations from $q$ source outputs and $1$ target output, in which only $q_1$ source outputs share information with the target output. For simplicity, the $q$ source outputs have equal  number of observations $n_1=...=n_q=n$, and the target output have less observations, i.e., $n_t < n$.  These observations form the training set and $n_{test}$ samples from the target output form the test set. 

For comparison, we take four other reference methods as benchmarks: 
\begin{enumerate}
\item
The non-regularized MGCP model, whose covariance structure is the same as the proposed model but without regularization, denoted as MGCP; 
\item
\textcolor{black}
{A regularized MGCP model with a full covariance structure, i.e., constructing the covariance among sources, denoted as MGCP-RF;}
\item
The two-stage method \cite{Kontar2020} denoted as BGCP-R, which first trains two-output GP models with regularization for each source and the target, then integrate the results of each sub-model in an empirical way;
\item
The single GP model constructed by a convolution process, in which only observations from the target output are used for training, denoted as GCP.
\end{enumerate}
\textcolor{black}
{In MGCP-RF, the sources and target are modeled as follows:
\begin{align}
y_i(\bm{x})&=g_{ii}(\bm{x})\ast Z_i(\bm{x})+g_{0i}(\bm{x})\ast Z_0(\bm{x})+\epsilon_i(\bm{x}), i \in \mathcal{I}^S \notag\\
y_t(\bm{x})&=\sum_{j \in \mathcal{I}}g_{jt}(\bm{x}) \ast Z_j(\bm{x})+g_{0t}(\bm{x}) \ast Z_0(\bm{x})+\epsilon_t(\bm{x}),
\end{align}
where $Z_0(\bm{x})$ is used for capturing shared information among sources, and $Z_i(\bm{x})$ is for unique information in each source/target. This structure refers to \cite{Kasarla2021}, but is tailored for transfer learning. To realize similar source-selection effect as MGCP-R, we penalize scale parameters both in $g_{0i}(\bm{x})$ and $g_{it}(\bm{x})$ as a group, i.e., $\mathbb{P}_{\gamma}(\bm{\theta}_0)=\gamma \sum_{i=1}^q \sqrt{\alpha_{0i}^2+\alpha_{it}^2}$. More details can be found in Appendix \ref{appendix:MGCP-RF}.}
In BGCP-R, $q$ regularized two-output GP models are trained using the data from each source and the target. The predictive distribution of each sub-model can be expressed as
\begin{align}
f_t(\bm{x}_*)| \bm{X}_{it}, \bm{y}_{it} \sim \mathcal{N} \left(\mu_i (\bm{x}_*), V_{if}(\bm{x}_*) \right), \notag
\end{align}
where $\bm{X}_{it}=(\bm{X}_i,\bm{X}_t)$, $\bm{y}_{it}=(\bm{y}_i^T,\bm{y}_t^T)^T$, $\mu_i (\bm{x}_*)=\bm{K}_*^T(\bm{X}_{it},\bm{x}_*)\bm{C}(\bm{X}_{it},\bm{X}_{it})^{-1}\bm{y}_{it}$, $V_{if}(\bm{x}_*)={\rm cov}_{tt}^f(\bm{x}_*,\bm{x}_*)-\bm{K}_*^T(\bm{X}_{it},\bm{x}_*)\bm{C}(\bm{X}_{it},\bm{X}_{it})^{-1} \bm{K}_*(\bm{X}_{it},\bm{x}_*)$. Then, the integrated results for BGCP-R is derived as
\begin{small}
\begin{align}
f_t(\bm{x}_*)| \bm{X}, \bm{y} \sim \mathcal{N} \left( \frac{\displaystyle \sum_{i=1}^{q} \mu_i  (\bm{x}_*) V_{if}^{-1}(\bm{x}_*)}{\displaystyle \sum_{i=1}^{q} V_{if}^{-1}(\bm{x}_*) }, \frac{q}{\displaystyle \sum_{i=1}^{q} V_{if}^{-1}(\bm{x}_*) }\right),
\label{eq:empirical combination}
\end{align}
\end{small}
which is an empirical combination of the predictions of each sub-model.
Gaussian kernel in \Eqref{eq:smooth kernel} is used for all methods and $L_1$ norm is used as the regularization function in MGCP-R and BGCP-R.

Regarding the model parameter settings, scaling parameters $\{\alpha_{ii},\alpha_{it}|i=1,...,q,t\}$ and noise parameter $\sigma$ for all outputs are initialized with random values in $[0,1]$. The length-scale diagonal matrix $\{\bm{\Lambda}_{ii},\bm{\Lambda}_{it}|i=1,...,q,t\}$ are also initialized with random values in $[0,1]$. Regarding the hyperparameter learning, we use L-BFGS method in GPflow \cite{GPflow2017}, which is a Python library based on TensorFlow, to maximize the log-likelihood. For the smoothing of $L_1$ norm regularization function, the value of parameter $\eta$ in \Eqref{eq:smooth approximation} is set to $10^{-5}$. 

Finally, to assess the prediction accuracy, we adopt the mean absolute error (MAE) criterion,
$${\rm MAE}=\frac{1}{n_{test}} \sum_{i=1}^{n_{test}}| f_t(\bm{x}_{*,i})-\hat{f}_t(\bm{x}_{*,i})|,$$
where $\hat{f}_t(\bm{x}_{*,i})$ is the predicted mean at $\bm{x}_{*,i}$. We repeat each case for $G=100$ times and present the distribution of the four methods' MAE in a group of boxplots.

\subsection{Simulation case \Romnum{1}}
\label{sec:4-simulation 1}
In order to assess the performance of different methods when the negative transfer of knowledge exists, i.e., learning some sources will bring negative influence on the learning of the target, we adopt an example with one-dimensional input.

The 1D example has $q=4$ source outputs defined in $\mathcal{X}_1=[0,5]$:
\begin{align*}
f_1(x)&=0.3(x-3)^3, & f_2(x)&=0.3x^2+2\sin(2x), \\
f_3(x)&=(x-2)^2, & f_4(x)&=(x-1)(x-2)(x-4) ,
\end{align*}
and one target output:
$$f_t(x)=0.2(x-3)^3+0.15x^2+\sin(2x).$$
The standard deviation of the measurement noise is set as $\sigma=0.2$. It can be found that the target output is a linear combination of the outputs $f_1$ and $f_2$. The other source outputs, which have different order ($f_3$) or zero points ($f_4$), are set as less-correlated sources.  The $n=30$ observations for each source are evenly spaced in $\mathcal{X}_1$, and $n_t=10$ observations for the target are evenly spaced in the left domain, $x \in [0,3]$. The $n_{test}=60$ test points are sampled uniformly in $\mathcal{X}_1$. Note that under such settings, the MAE at these test points contains both the interpolation error and the extrapolation error. 

\begin{figure}[!t]
\centering
\includegraphics[width=3.0in]{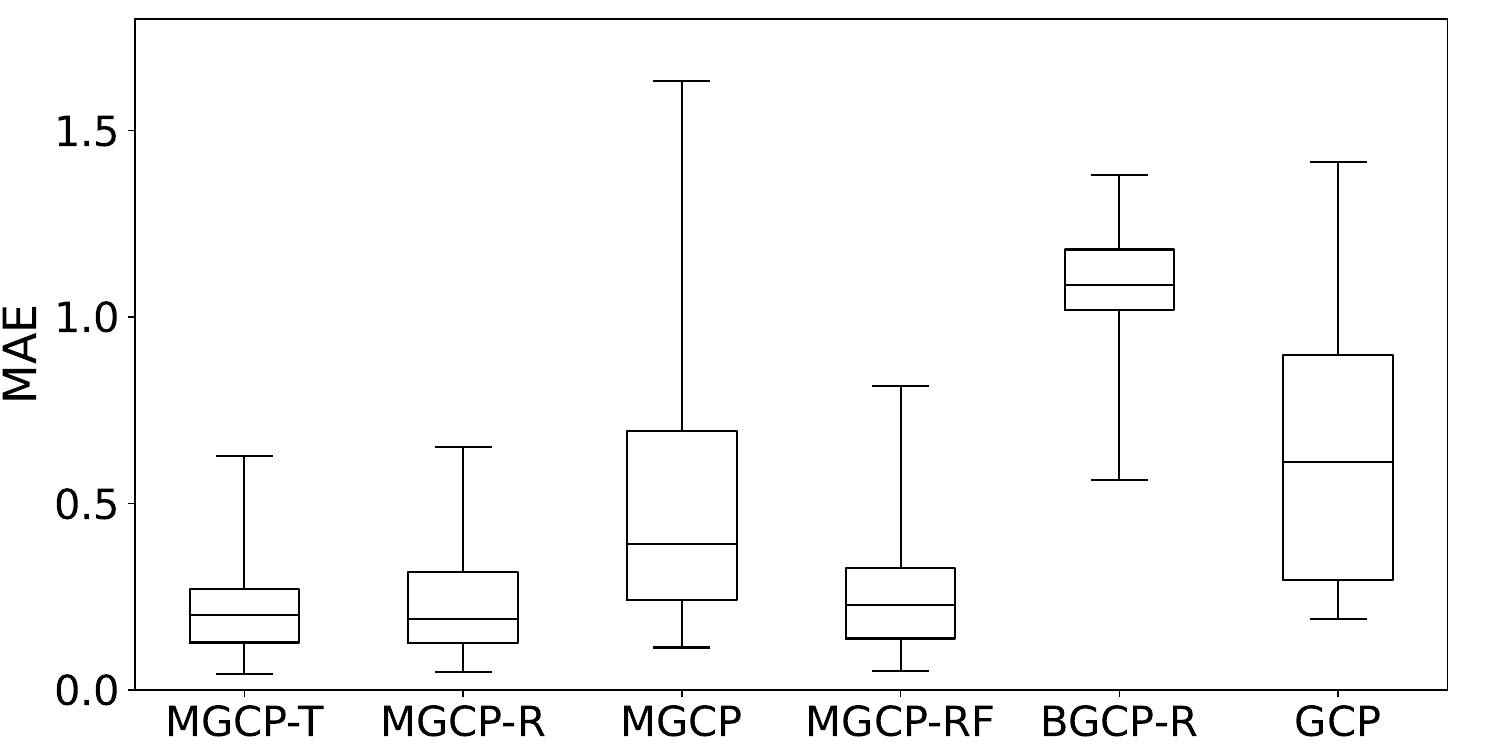}
\caption{Boxplots of the MAE in the simulation case I, where the line in each box represents the median value.}
\label{fig:case1 boxplot}
\end{figure}
\begin{figure*}[!t]
\centering
\includegraphics[width=6in]{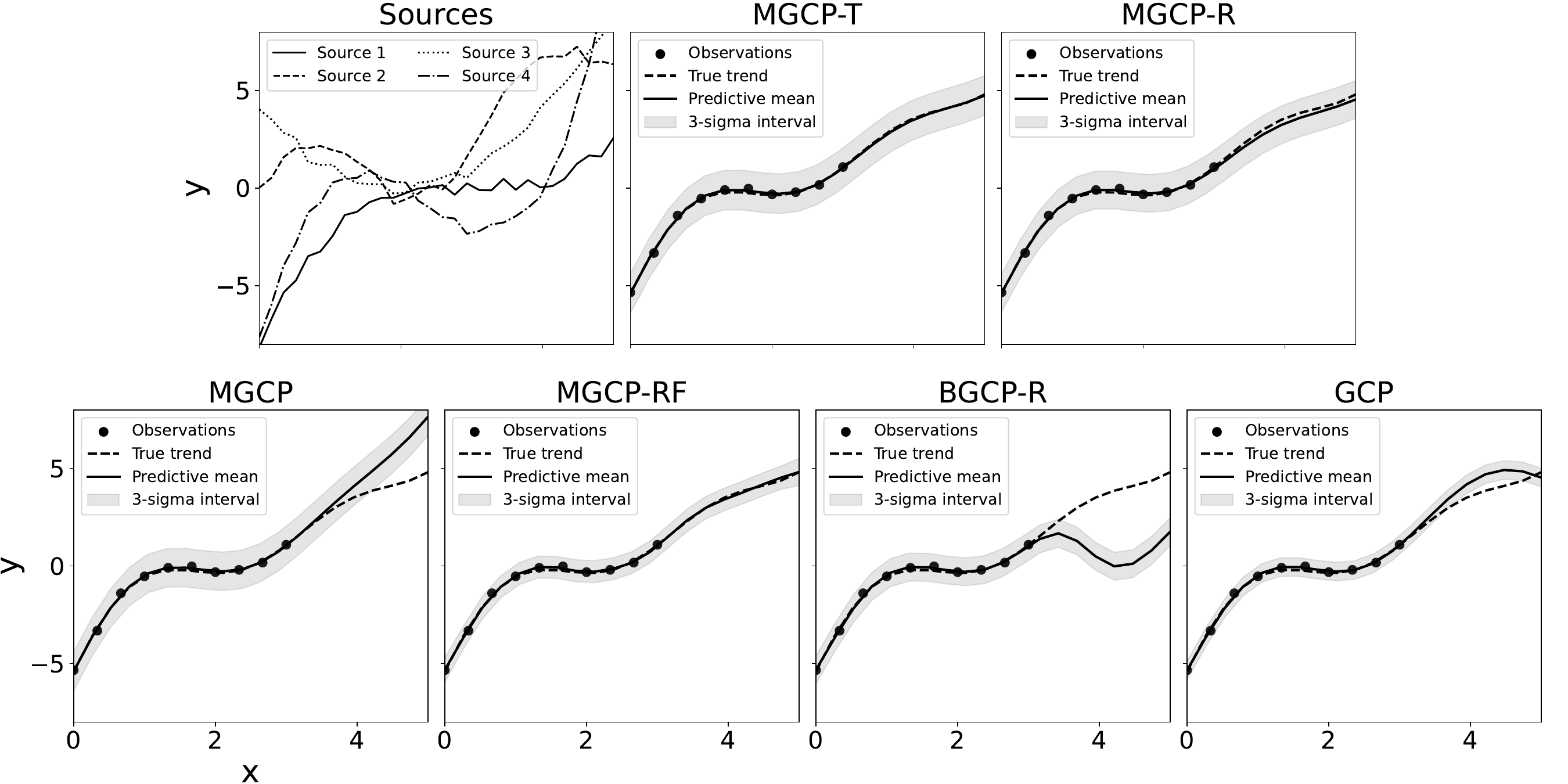}
\caption{Visualization of results in one repetition of 1D example.}
\label{fig:case1 1D results}
\end{figure*}

Considering that the target is a combination of two sources, we benchmark with another method denoted as MGCP-T, which only uses the source outputs $f_1$ and $f_2$ to construct a non-regularized MGCP model. MGCP-T is set as the underlying true model and possesses the true covariance structure, wherein negative transfer will not happen. It presents the optimal predictive performance in all introduced methods. 
Figure \ref{fig:case1 boxplot} shows the boxplots of MAE in these two examples, and \Figref{fig:case1 1D results} shows the data of each source and the predicted trends of the target in one repetition of the 1D example.


Firstly, we focus on three methods, MGCP-R, MGCP-T and MGCP.
The results shown in \Figref{fig:case1 boxplot} illustrate the superior performance of our method. 
MGCP-R performs similarly with MGCP-T and provides much more accurate and stable prediction than MGCP. Note that MGCP-T is the true model with the smallest median and variance value of MAE. This result exactly verifies the conclusion claimed in Section \ref{sec:3-statistical} that our regularized model possesses the ability of selecting informative sources. The negative transfer of information caused by $f_3$ and $f_4$ is greatly reduced in the proposed method. 
To state the above conclusion more clearly, we compare part of the estimated parameters of MGCP-T, MGCP-R and MGCP in one repetition of the 1D example. \autoref{tab:estimated parameters} shows the parameters belonging to the smooth kernels $g_{it},i\in \mathcal{I}^S$, which connect the target and each source. As shown in the table, three methods provide similar estimators except the scaling parameters in $g_{3t}$ and $g_{4t}$, which are shrunk to nearly zero in MGCP-R but not in MGCP. We can also directly observe from \Figref{fig:case1 1D results}, a visualization of \autoref{tab:estimated parameters}, that the predictive mean of the target by MGCP presents an obvious linear shift-up in the right domain, and the sources $f_3$ and $f_4$ also shift up linearly at the same area. Moreover, we would like to mention that our method is robust towards both the linear correlation and the non-linear correlation. As the correlation analysis shown in \autoref{tab:correlation in case I}, in the 1D example, the Pearson correlation between $f_4$ and $f_t$ is very high while the Kendall correlation between them is low. MGCP-R is not misled by the high linear correlation between $f_4$ and $f_t$ since it can comprehensively consider both the linear and non-linear relationships and their combinations.
In addition, all source outputs are correlated with each other, which is not listed in \autoref{tab:correlation in case I}.

\begin{table}[!t] 
\renewcommand{\arraystretch}{1.3}
\setlength\tabcolsep{4pt}
\caption{Part of the estimated parameters in one repetition of the 1D Example}
\label{tab:estimated parameters}
\centering
\begin{tabular}{l c cccc}
\hline
\multicolumn{2}{c}{Methods}&MGCP-T  & MGCP-R & MGCP & MGCP-RF\\
\hline
 & $\alpha_{1t}$ & 14.92 & 12.68 & 12.62  &2.83\\
 Scaling & $\alpha_{2t}$ & 2.15 & 1.58 & 2.04 & 1.56\\
 parameters & $\alpha_{3t}$  & - & \textbf{1.00e-5} & 2.53 & 7.00e-6\\
 & $\alpha_{4t}$ & - & \textbf{1.11e-4} & 3.96 &8.00e-2\\
 \hline
 & $\bm{\Lambda}_{1t}$ & 2.49 & 2.40 & 2.62 & 1.06\\
 Length-scale& $\bm{\Lambda}_{2t}$ & 0.87 & 0.81 & 0.89 &0.83\\
 parameters & $\bm{\Lambda}_{3t}$ & - & 3.72 & 4.07 &1.17\\
 & $\bm{\Lambda}_{4t}$ & - & 8.56 & 1.50 & 2.97\\
\hline
\end{tabular}
\end{table}

\begin{table}[!t] 
\renewcommand{\arraystretch}{1.3}
\setlength\tabcolsep{4pt}
\caption{Correlation between each source and the target in simulation case I}
\label{tab:correlation in case I}
\centering
\begin{tabular}{c c cccc}
\hline
 ~& Type & $f_1:f_t$  & $f_2:f_t$ & $f_3:f_t$ &$f_4:f_t$ \\
\hline
\multirow{2}{*}{1D example} & Pearson &0.895 &0.840 &0.555 &0.745  \\
& Kendall & 0.922 & 0.559 & 0.416 & 0.443 \\
\hline
\end{tabular}
\end{table}

\textcolor{black}
{An interesting observation is that the MGCP-RF performs only comparable with the MGCP-R in \Figref{fig:case1 boxplot}, although it predicts a little more accurately and also selects the two informative sources in one repetition presented in \autoref{tab:estimated parameters} and \Figref{fig:case1 1D results}. 
We believe this is due to its larger parameter space. Under same circumstances, MGCP-RF needs $50\%$ more parameters to construct, which poses great challenge in parameter estimation. The following analysis on higher dimension inputs and/or outputs in Sections \ref{sec:4-simulation 2} and \ref{sec:4-simulation 3} also confirms our findings. }

The median of prediction error of BGCP-R is the largest. This is because BGCP-R focuses on the  information transfer from each individual source pair-wisely and cannot incorporate the information of all sources globally. As a result, the negative transfer happens to BGCP-R when the target contains combination of sources, i.e., which leads to larger prediction error than only using the target data (GCP). This is one of major shortcomings of BGCP-R since there are very rare cases in practice that the target and source share the same functional form. From the results in \Figref{fig:case1 1D results}, we can observe this influence more clearly: the predictive mean of BGCP-R has a similar valley shape with $f_1$ in the right domain. 

\textcolor{black}{
Finally, the influence of the tuning parameter $\gamma$ is worth attention, which serves as a similar role of the tuning parameter in LASSO, i.e., there will be a continual selection path as we increase the value of $\gamma$. The larger $\gamma$ is, the less sources will be selected, which means too-large $\gamma$ may bring negative influence due to the exclusion of some relatively-weak-informative sources. To demonstrate this, we apply MGCP-R to model $f_1, f_2$ and $f_t$ with varying values of $\gamma$. More details and experiment results can be found in Appendix \ref{appendix:tuning-parameter}.}

\subsection{Simulation case \Romnum{2}}
\label{sec:4-simulation 2}
In this subsection, we apply the proposed framework to transfer information from the source with  inconsistent input domain to the target. We adopt $p=3$ source outputs:
\begin{align*}
f_1(x_1)&=3\sin(x_1), \\ 
f_2(x_1,x_2)&=4\cos(2x_1)+x_2^2+x_2, \\ 
f_3(x_1,x_2)&=2\sin(2x_1)+x_2^2,
\end{align*}
and one target output:
\begin{align*}
f_t(x_1,x_2)=2\sin (x_1)+x_2^2+x_2.
\end{align*}
where $x_1 \in \mathcal{X}_1 = [-2,2]$ and $x_2 \in \mathcal{X}_2 = [-2,2]$. The standard deviation of measurement noise is also set as $\sigma=0.2$. In this case, the source $f_1$ has the inconsistent input domain with the target. Besides, $f_1$ is set as the obtained mean of marginal distribution after our domain adaptation method. 

According to the notations in Section \ref{sec:3-inconsistent}, for the source $f_1$, the common input feature is $\bm{x}^{(c)}=x_1$ and the unique input feature is $\bm{x}^{(t)}=x_2$. Thus, following the procedure of DAME, we firstly generate $n_{1^{\prime}}=8$  induced data for $f_1$, $\{\bm{x}^{(c)}_{1^{\prime},a}, y_{1^{\prime},a}\}_{a=1}^{8}$ evenly spaced in $\mathcal{X}_1$. Then, choose another eight points $\{\bm{x}^{(t)}_{1^{\prime \prime},b}\}_{b=1}^{8}$ evenly spaced in $\mathcal{X}_2$. 
The 64 pseudo data of the source $f_1$ can be obtained through \Eqref{eq:pseudo data}, where $\epsilon_{a,b}\sim \mathcal{N}(0, 0.2^2)$ is the same as the measurement noise of target.  
For the other two sources, $n=64$ sample points are generated at the same location in $\mathcal{X}_1 \times \mathcal{X}_2$. 
For the target oputput, $n_{t}=24$ sample points are located at the nodes of a $3 \times 8$ grid in $[0,2]\times[-2,2]$, and $n_{test}=100$ test points are uniformly spaced in $\mathcal{X}_1 \times \mathcal{X}_2$. 
In order to identify the effect of our domain adaptation method, the first source's data are not used in MGCP and BGCP-R.

\begin{figure}[!t]
\centering
\includegraphics[width=2.8in]{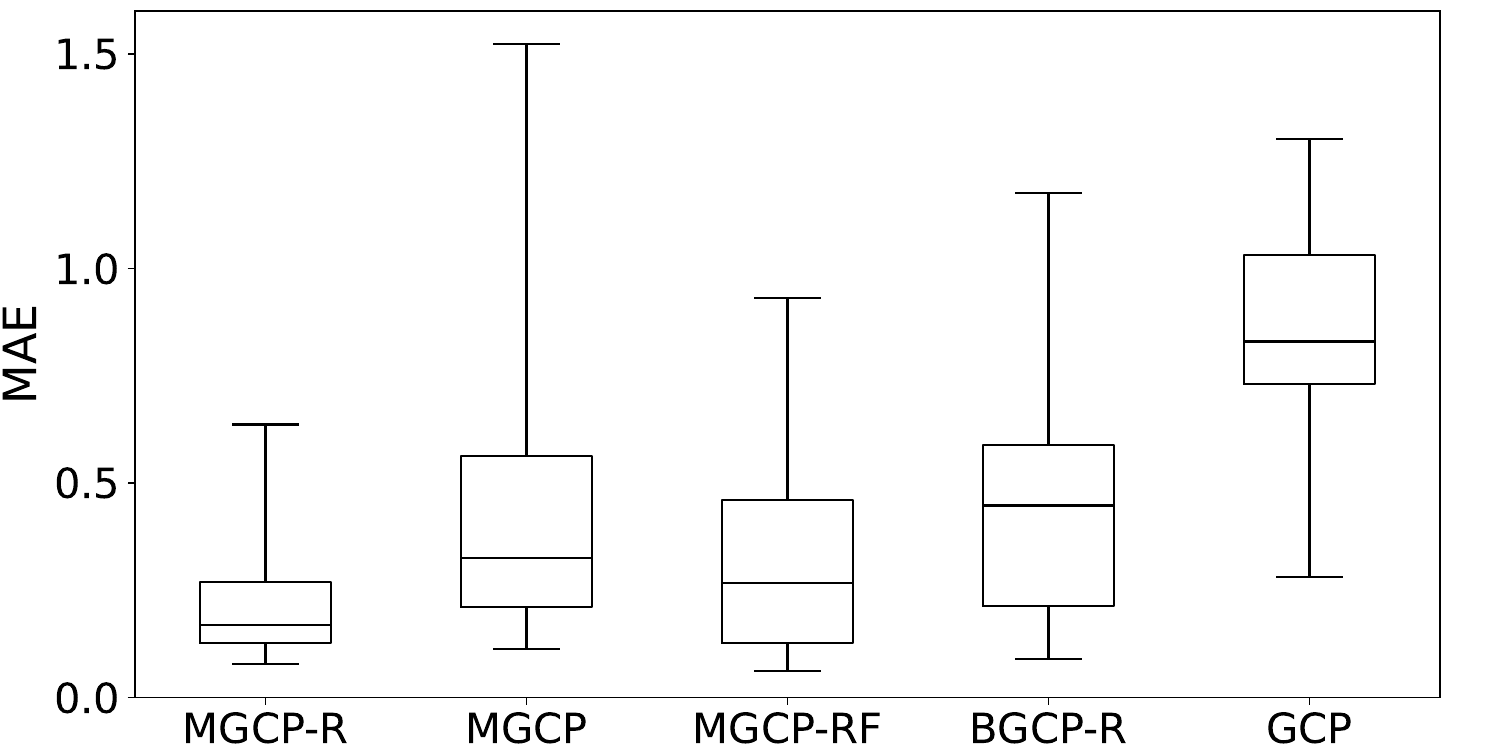}
\caption{Boxplot of MAE in the simulation case \Romnum{2}. The line in each box represents the median value.}
\label{fig:case2 boxplot}
\end{figure}

\begin{figure}[!t]
\centering
\includegraphics[width=2.8in]{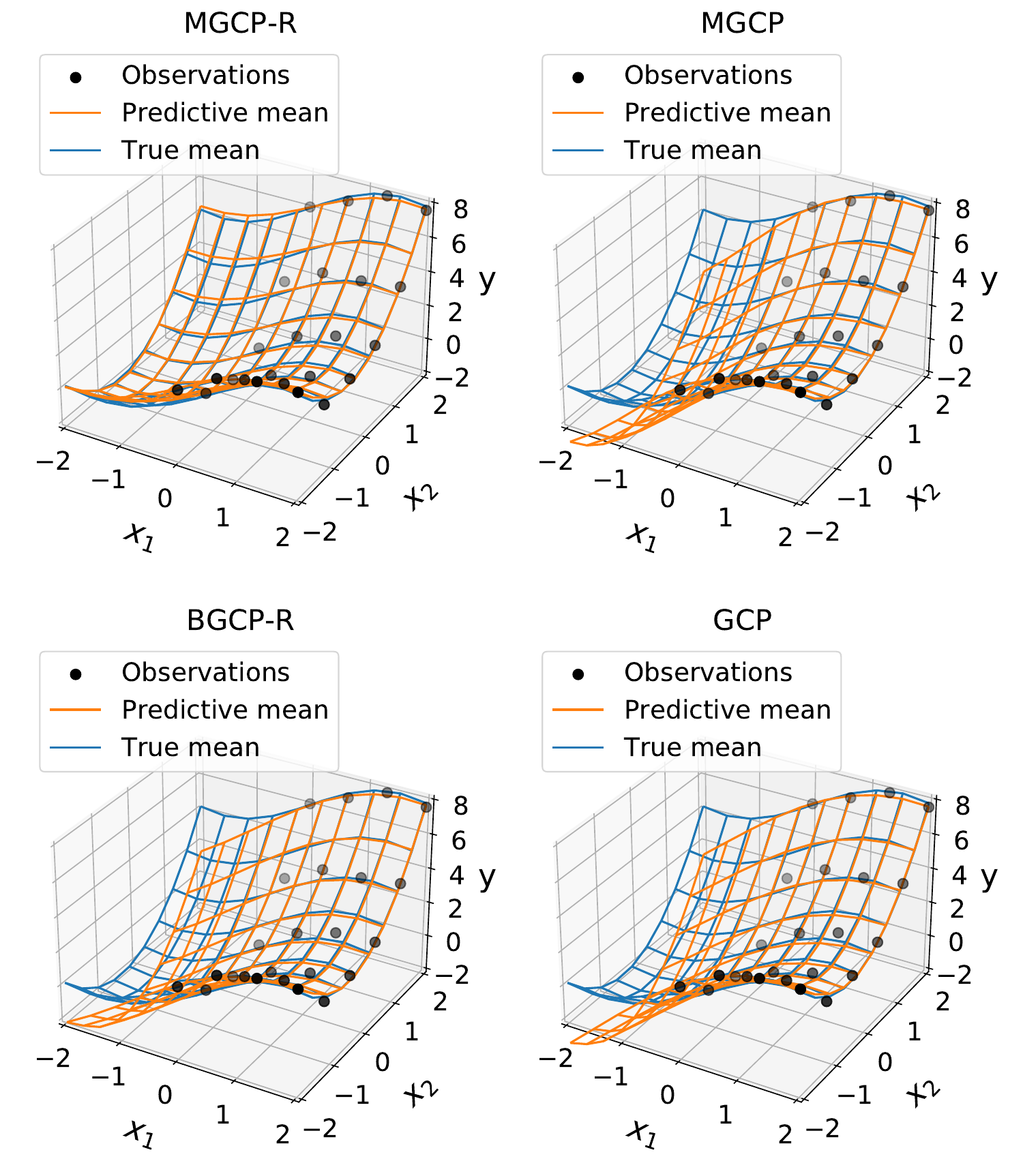}
\caption{Predictive results in one repetation of the simulation case \Romnum{2}.}
\label{fig:case2 results}
\end{figure}

The results shown in \Figref{fig:case2 boxplot} and \Figref{fig:case2 results} demonstrate the effectiveness of our modeling framework, especially the DAME approach. As the observations of the target are located in the half domain $[0,2]\times[-2,2]$, the information of the target's behavior along $x_1$ can only be borrowed from the source $f_1$, which leads to the superior performance of the proposed method in the boxplot of MAE. 
Predictive results of one repetition in \Figref{fig:case2 results} also verify the above conclusion, where we can clearly see that MGCP-R is the only method providing accurate fitting in both $x_1$ and $x_2$ directions. 
Besides, as shown in the boxplot, the predictive accuracy of MGCP and BGCP-R are better than GCP, because $f_2$ and $f_3$ can also provide some beneficial information to the target prediction. However, as more informative information along $x_1$ is contained in the first source, our method can effectively leverage this knowledge and predict the target more accurately. 

\color{black}
\subsection{Simulation case III}
\label{sec:4-simulation 3}
The above simulation cases have demonstrated the effectiveness of our method with small number of sources and input dimensions. In this section, we aim to verify the performance of MGCP-R with more sources and higher input dimensions. 

Setting 1: To test the performance with more sources, we adopt similar setting in the 1D example of Section \ref{sec:4-simulation 1} and define the following four kinds of functions:
\begin{align*}
f_{k}(x)&=0.3(x-2.5-e_1^k)^3, \\
f_{n_e+k}(x)&=0.3x^2+2\sin(2x+e_2^k), \\
f_{2n_e+k}(x)&=(x-1.5-e_3^k)^2, \\
f_{3n_e+k}(x)&=(x-1)(x-2)(x-3.5-e_4^k) ,
\end{align*}
where $n_e$ is the number for each kind of sources, $e_i^k$ are uniformly sampled from $[0,1]$, and $k=\{1,...,n_e\}$. 
We define the target output as:
$$f_t(x)=0.2(x-2.5-e_1^1)^3+0.15x^2+\sin(2x+e_2^1).$$
In this setting, we take $n_e = 2,4,10$, thus the maximum  number of outputs are $41$ (including the target). We keep the other settings same as simulation case I and repeat the experiments $50$ times.

Setting 2: Regarding the ability of our method with higher input dimensions, we define the following three kinds of sources:
\begin{align*}
f_{k}(\bm{x})&=3\sum_{j=1}^2\sin(x_j+e_{1j}^k) , \\
f_{n_e+k}(\bm{x})&=4\sum_{j=1}^2\cos(2x_j+e_{2j}^k)+x_3^2+x_3+2x_4-x_5, \\
f_{2n_e+k}(\bm{x})&=2\sum_{j=1}^2\sin[2(x_1+e_{3j}^k)]+x_3^2-x_4+2x_5, 
\end{align*}
where $e_{ij}^k$ are uniformly sampled from $[-0.25,0.25]$. 
The target output is:
$$f_t(x)=2\sum_{j=1}^2\sin(x_j+e_{1j}^1)+x_3^2+x_3+2x_4-x_5.$$
In this setting, we take $n_e = 1,2$, thus the maximum number of outputs are $7$. The number of input dimension is $5$ and the inconsistent dimensions are $3$. 
Similarly to the simulation case II, $\{f_{i}(\bm{x})\}_{i=1}^{n_e}$ is set as the mean of two-dimensional marginal distribution. To apply the  domain adaptation method to these sources, we first generate $n_{1^{\prime}}=10$ induced data in $\mathcal{X}_1=\{x_1,x_2\}$ from $\bm{x}\sim \mathcal{N}(\bm{0},\bm{I}_2)$. Then, we randomly choose $10$ points in $\mathcal{X}_2=\{x_3,x_4,x_5\}$ ($\bm{x}\sim \mathcal{N}(\bm{0},\bm{I}_3)$) for each induced data to generate 100 pseudo data. For the other sources, $n=100$ observations for each source are sampled randomly from $\bm{x}\sim \mathcal{N}(\bm{0},\bm{I}_5)$. 
For the target, $150$ samples are generated in the same way and $50$ of them, which satisfies $x_1>0$, are picked as training data. Then, another $150$ samples are randomly generated as test data. 

The average prediction error shown in \autoref{tab:case3 MAE} reveals that MGCP-R still performs the best under more sources and higher input dimensions. For a moderate number of sources (40 when $n_e=10$), MGCP has severe negative transfer effect compared with GCP. As $n_e$ increases,  the difference between MGCP-RF and MGCP-R gets larger, which is expected due to the higher parameter space of MGCP-RF, and thus a large number of hyper-parameters to be optimized.

We also provide the average optimization and prediction time in \autoref{tab:case3 time} for one random start (five random starts in one repetition). The computational load of MGCP-RF is much heavier than other methods, which is a severe drawback. Comparing MGCP-R and MGCP, their prediction time is close but the former's optimization time is less, which is another advantage of regularization. For BGCP-R, it needs less optimization time than MGCP-R in setting 1 due to its smaller parameter space, which makes local optima more easy to reach. However, BGCP-R's optimization time in setting 2 is more than MGCP-R. This is because the difference between their parameter dimensions is smaller than setting 1, and inversing the covariance matrix ($O(qn^3+n_t^3)$ for MGCP-R, $O(q(n+n_t)^3)$ for BGCP) in optimization dominates the computational complexity.

\begin{table}[b] \color{black}
\renewcommand{\arraystretch}{1.2}
\setlength\tabcolsep{3pt}
\setlength\abovecaptionskip{0cm}  
\setlength\belowcaptionskip{0cm} 
\caption{Average MAE of each method in simulation case III.}
\label{tab:case3 MAE}
\centering
\begin{small}
\begin{tabular}{c l| ccccc}
\hline
 Setting &\makecell *[c]{outputs} & \makecell*[c]{MGCP\\-R} &  \makecell*[c]{MGCP\\ }&  \makecell*[c]{MGCP\\-RF} & \makecell*[c]{BGCP\\-R}  &  \makecell*[c]{GCP\\} \\
\hline
 \multirow{3}{*}{1} & 9\ \ ($n_e=2$)  & 0.251 & 0.543 & 0.547 & 0.539 & 0.609  \\
& 17($n_e=4$)  & 0.323 &0.865 &0.830&0.682  &0.735  \\
& 41($n_e=10$)   & 0.273 & 1.211 &0.667 &  0.495  & 0.652 \\
\hline
 \multirow{2}{*}{2} & 4\ \ ($n_e=1$)  & 0.349 & 0.575& 0.361 & 0.631  & 0.764  \\
 & 7\ \ ($n_e=2$)  &0.377  &0.556& 0.601 & 0.612  & 0.755 \\
\hline
\end{tabular}
\end{small}
\end{table}
\begin{table}[h] \color{black}
\renewcommand{\arraystretch}{1.2}
\setlength\tabcolsep{3pt}
\setlength\abovecaptionskip{0cm}  
\setlength\belowcaptionskip{-0.2cm} 
\caption{Average optimization and prediction time (value in the parentheses) of each method in simulation case III.}
\label{tab:case3 time}
\centering
\begin{small}
\begin{tabular}{c l| ccccc}
\hline
Setting &\makecell *[c]{outputs} & \makecell*[c]{MGCP\\-R} &  \makecell*[c]{MGCP\\ }  &  \makecell*[c]{MGCP\\-RF} & \makecell*[c]{BGCP\\-R} &  \makecell*[c]{GCP\\} \\
\hline
 \multirow{6}{*}{1}&\multirow{2}{*}{9\ \ $(n_e=2) $} & 8.86 & 19.89& 35.82 & 7.72  & 0.26  \\
&& (0.16) & (0.16)& (0.27) & (0.18)  & (0.01) \\ \cline{3-7}
&\multirow{2}{*}{17$(n_e=4) $} & 16.57 &50.90 &176.93&15.48 &0.21  \\
&& (0.23) & (0.23)& (0.79) & (0.37)  & (0.01) \\ \cline{3-7}
&\multirow{2}{*}{41$(n_e=10) $} & 62.90 &281.29 &3673.4 &39.47 &0.22  \\
&& (0.59) & (0.58) & (4.43) & (0.91) & (0.01) \\ 
\hline
 \multirow{4}{*}{2}&\multirow{2}{*}{4\ \ $(n_e=1) $} & 34.26 & 109.46& 328.22 & 40.71  & 0.28  \\
&& (0.08) & (0.09) & (0.12)& (0.12)  & (0.01) \\ \cline{3-7}
&\multirow{2}{*}{7\ \ $(n_e=2) $} & 31.24 & 372.14 & 1831.37 & 85.04  & 0.30  \\
&& (0.15) & (0.15) & (0.31) & (0.24) & (0.01) \\
\hline
\end{tabular}
\end{small}
\end{table}

\color{black}
\subsection{Real case of ceramic manufacturing}
\label{sec:4-real case}
In this real case study, the goal is to predict the response surface of ceramic product's density. 

\subsubsection{Data description}
The data are collected through two groups of experiments differing in manufacturing methods and process parameters. 
The first group contains 28 ($4 \times 7$) experiments using dry pressing manufacturing technique under 4 pressures and 7 temperatures. 
The second group contains 16 ($4 \times 4$)  experiments using stereolithography-based additive manufacturing under 4 solids loadings and 4 temperatures. 
Table \ref{tab:controlled parameters} summarizes the values of controlled parameters and all other process parameters are kept fixed in each group. 
As only the temperature is the shared input parameter, the domain adaptation is needed for leveraging information from the data of one manufacturing method to the other.

\begin{figure}[!t]
\centering
\includegraphics[width=2.8in]{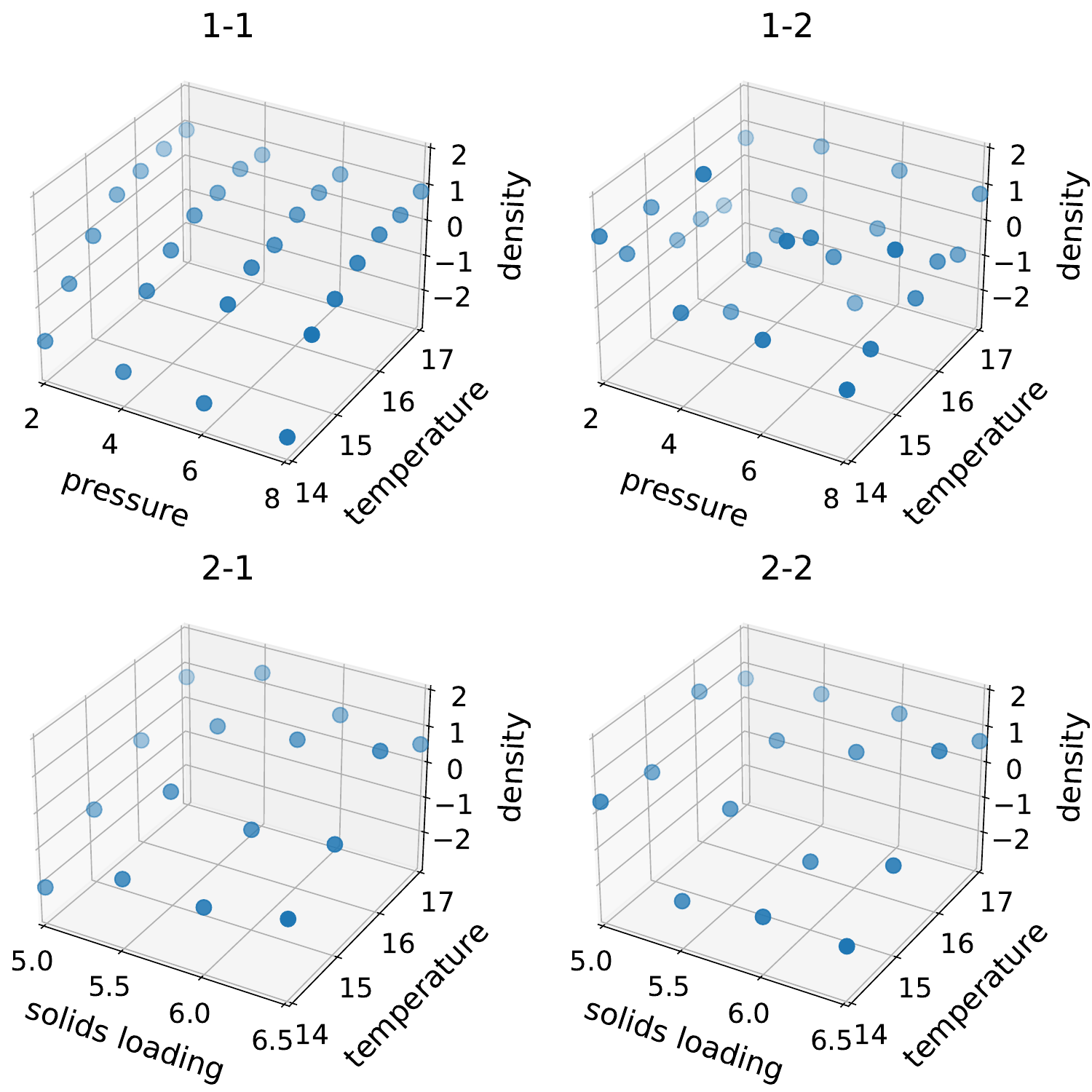}
\caption{Data for the ceramic manufacturing. 1-1, 1-2 are from dry pressing manufacturing and 2-1, 2-2 are from additive manufacturing, where the second index represents the measurement method: 1 for mass-volume method and 2 for Archimedes method.}
\label{fig:real case data}
\end{figure}

Two methods, mass-volume method and Archimedes method, are used to measure the density, so there are two sets of measurement for each group. 
The overall 4 datasets are shown in \Figref{fig:real case data}, where the first index of dataset represents the manufacturing method and the second index represents the measurement method. 
Density data of each dataset are standardized to have zero mean and unit variance. 
Note that for the same group of experiments, two measurement methods give different response surfaces because the size of ceramic product is small. 
In this case, the measurement error of Archimedes method might be higher, resulting in negative transfer if we incorporate it in the transfer learning.  
We treat the density data of 1-1 as the target output and the remaining 3 datasets as source outputs. 
For the target, only 8 data points are randomly chosen as observations and the rest 20 points are used for testing.

For MGCP and BGCP-R, only 1-2, which have the same input domain as the target, is used as the source data in the model. In such condition, MGCP degenerates to a two-output Gaussian Process model, so the main difference between it and BGCP-R is that the regularization in the latter model provides the ability to reduce the negative transfer of knowledge. 
For our method, we apply DAME to the sources 2-1 and 2-2 as follows. Firstly marginalize  the original data to the input domain only with `temperature' feature. Then conduct kernel regression to obtain the mean of marginal distribution.  In this case, we use 7 induced points to keep up with the number of target data. Then expand them to the target input domain and obtain 28 pseudo data. Note that the adaptation process of 2-1 has been shown in \Figref{fig:domain adaptation} before. Thus, we have equal number of data for the original dataset 1-2 and the pseudo dataset of 2-1, 2-2, i.e., $n_{1-2}=n_{2-1}=n_{2-2}=28$. Finally, they are taken as 3 source outputs to establish a regularized MGCP model, where $L_1$ norm regularization is implemented. 

\begin{table}[!t] 
\renewcommand{\arraystretch}{1.3}
\caption{Controlled parameters in ceramic product manufacturing}
\label{tab:controlled parameters}
\centering
\begin{tabular}{ccc}
\hline
Parameter &\makecell *[c] {Dry pressing} & \makecell *[c]{Additive\\ manufacturing} \\
\hline
 Temperature($\times$ 100 \textcelsius) &  \makecell*[c]{14, 14.5, 15, 15.5,\\ 16, 16.5, 17}&  14, 15, 16, 17  \\
 Pressure($\times 10^8$ Pa)  & 2, 4, 6, 8 &-  \\
 Solids loading($\times$10\%) &- & 5, 5.5, 6, 6.5 \\
\hline
\end{tabular}
\end{table}

\subsubsection{Performance Evaluation}
Firstly we provide some intuitive understanding of the advantage of our method. 
From the experimental data, it can be found that temperature is the key factor affecting the density of ceramic products. 
The trend of density in 1-1 is similar to that in 2-1, which makes it feasible to transfer information from 2-1 to 1-1, i.e., from one manufacturing method to another. 
This application is highly desirable in real world as the cost of lab experiment is highly expensive. 
For example, each data point in \Figref{fig:real case data} takes 20 hours to produce. 
Borrowing knowledge from previous research or experiments can greatly reduce the generation of new samples, and acquire a more accurate response surface efficiently and cheaply.

\begin{figure}[!t]
\centering
\includegraphics[width=3.0in]{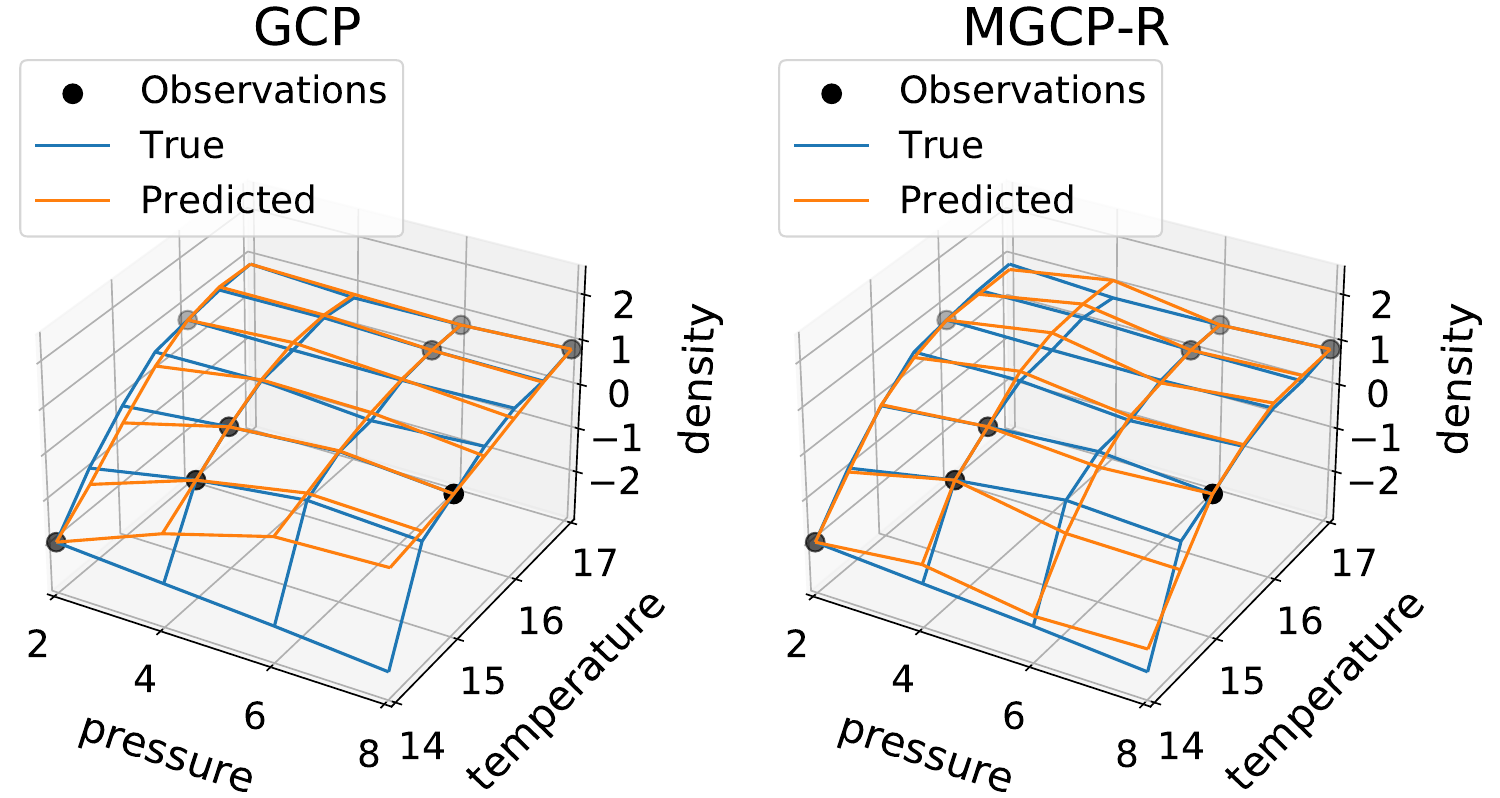}
\caption{Prediction results in one repetition of the ceramic manufacturing case.}
\label{fig:caseReal results}
\end{figure}

We repeat the case 50 times and the results are shown in Table \ref{fig:caseReal MAE}. 
The mean and variance of MAE illustrates that MGCP-R outperforms the other benchmarks, with the help of regularization and data from other manufacturing method. 
\textcolor{black}{
The results of full-covariance method MGCP-RF are close to MGCP-R, as the optimization problem in large parameter space is not severe for MGCP-RF with a small number of data. Nevertheless, the larger variance of MGCP-RF comparing to MGCP-R and the lower computational complexity of MGCP-R still demonstrate the superiority of our proposed method.}
The performance of BGCP-R is almost the same as GCP, while MGCP performs worst in all methods. 
This suggests that the information transferred from the source 1-2 misleads the prediction for the target in MGCP, but BGCP-R reduces its influence to nearly zero through regularization. 
From the predictive results in \Figref{fig:caseReal results}, we can clearly see that MGCP-R is capable of recovering the response surface more accurately with only a few experimental samples, when some historical sources can offer some informative information. 

\begin{table}[!t] 
\renewcommand{\arraystretch}{1.3}
\caption{Prediction error of each method in the ceramic manufacturing case}
\label{tab:error in real case}
\centering
\begin{tabular}{c c cccc}
\hline
   & MGCP-R  & MGCP & MGCP-RF & BGCP-R &GCP \\
\hline
 Mean &0.286 &0.398 & 0.288 &0.340 &0.362  \\
 Std. & 0.106 & 0.201 & 0.121 & 0.172 &0.235 \\
\hline
\end{tabular}
\label{fig:caseReal MAE}
\end{table}

\section{Conclusion}
\label{sec:5-conclusion}
We propose a regularized MGCP modeling framework that can select informative source outputs globally and transfer information from sources with both consistent input domain and inconsistent input domains. 
Our work firstly conducts convolution process to establish a special covariance structure that models the similarity within and across outputs. Then, a regularized maximum log-likelihood estimation is performed based on the structure. 
Some statistical properties are also derived to guarantee the effectiveness of our method. 
A domain adaptation approach based on marginalization and expansion deals with the inconsistent input domain of sources successfully. 
Both simulation cases and the real case of ceramic manufacturing demonstrate the superiority of our method.

There are several open topics worthy of investigation in the future based on our work. 
The first one is to apply our method to classification problems, where the posterior distribution needs to be approximated as it doesn't have an explicit form. 
One important issue of classification problems is that the data usually contain considerable amount of features, e.g., gene expression, which increases the complexity and computational burden of GP model.
Therefore, the selection of informative sources and critical features should be combined together, and computationally-efficient algorithms are needed to train the model with high-dimensional data. 
The second one is considering the correlated noise.
For example, in time series analysis, the auto-correlated noise should be considered, which may greatly increase the prediction accuracy and improve the flexibility of the MGCP model. 
Thirdly, in the proposed approach, MGCP modeling and domain adaptation are treated as two separate tasks. Jointly optimizing these two tasks in a unified framework will be studied in future.

\ifCLASSOPTIONcompsoc
  \section*{Acknowledgments}
\else
  \section*{Acknowledgment}
\fi

\ifCLASSOPTIONcaptionsoff
  \newpage
\fi

\begin{IEEEbiography}[{\includegraphics[width=1in,height=1.25in,clip,keepaspectratio]{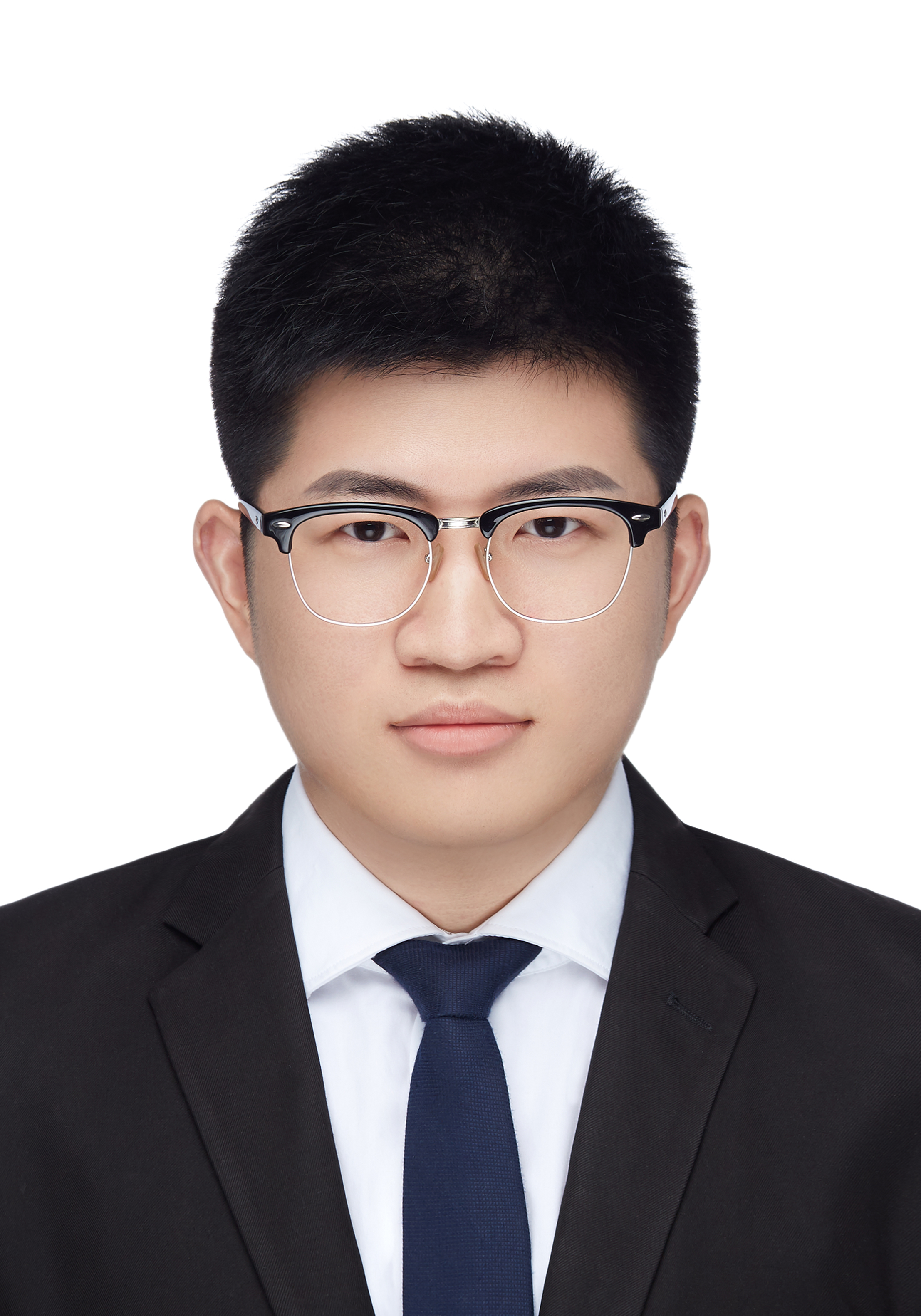}}]{Xinming Wang}
received the B.S. degree in Mechanical Engineering from Tsinghua University, Beijing, China, in 2020. He is currently working towards the Ph.D. degree in Industrial and System Engineering with Peking University, Beijing, China. His current research interests include data science,  transfer learning, and intelligent manufacturing.
\end{IEEEbiography}
\begin{IEEEbiography}[{\includegraphics[width=1in,height=1.25in,clip,keepaspectratio]{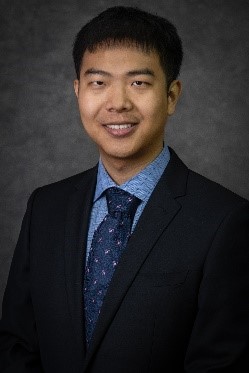}}]{Chao Wang}
is an Assistant Professor in the Department of Industrial and Systems Engineering at the University of Iowa. He received his B.S. from the Hefei University of Technology in 2012, and M.S. from the University of Science and Technology of China in 2015, both in Mechanical Engineering, and his M.S. in Statistics and Ph.D. in Industrial and Systems Engineering from the University of Wisconsin-Madison in 2018 and 2019, respectively. His research interests include statistical modeling, analysis, monitoring and control for complex systems. He is member of INFORMS, IISE, and SME.
\end{IEEEbiography}
\begin{IEEEbiography}[{\includegraphics[width=1in,height=1.25in,clip,keepaspectratio]{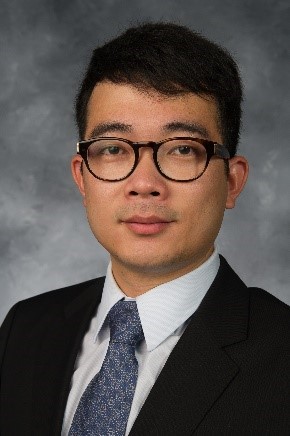}}]{Xuan Song}
is an assistant professor at the department of industrial and systems engineering at the University of Iowa. His research interest is additive manufacturing process development and optimization as well as novel applications of AM technologies in various areas, such as biomedical imaging, tissue engineering, energy harvest, robotics, etc. At UIowa, Dr. Song’s research focuses on the development of next- generation additive manufacturing processes with multi-material, multi-scale or multi-directional capabilities. He obtained his Ph.D. degree in industrial and systems engineering from the University of Southern California in 2016.
\end{IEEEbiography}
\begin{IEEEbiography}[{\includegraphics[width=1in,height=1.25in,clip,keepaspectratio]{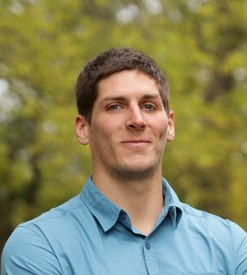}}]{Levi Kirby}
is a PhD student at the University of Iowa. He obtained his Bachelor’s and Master’s Degree from Western Illinois University in engineering technology. At Iowa, his research focuses on various forms of additive manufacturing, including printing of energetic composites and highly dense ceramics. Throughout his collegiate career, he has been awarded the E. Wayne Kay Scholarship, the Departmental Scholar, Magna Cum Laude, and 3MT finalist.
\end{IEEEbiography}
\begin{IEEEbiography}[{\includegraphics[width=1in,height=1.25in,clip,keepaspectratio]{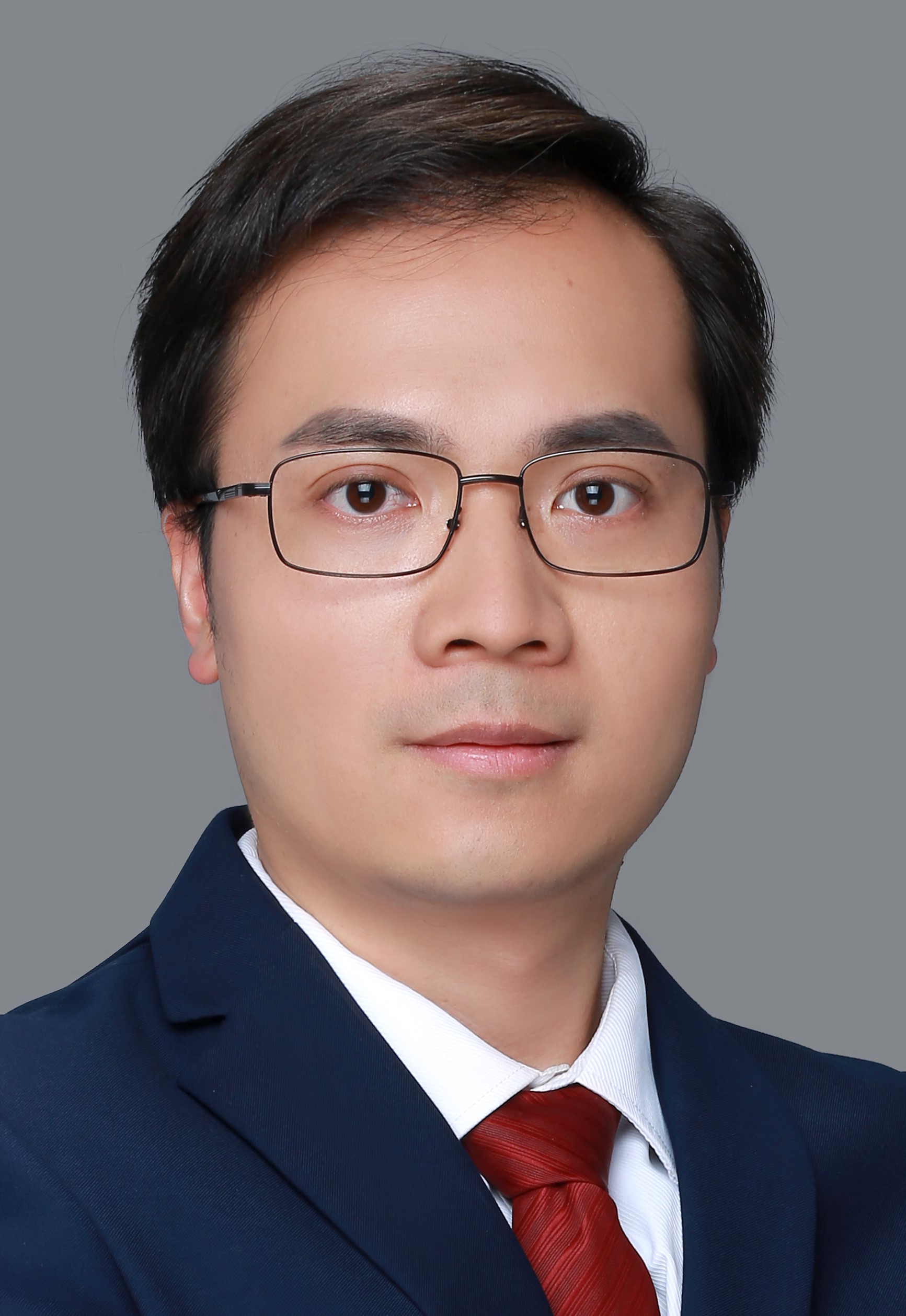}}]{Jianguo Wu}
received the B.S. degree in Mechanical Engineering from Tsinghua University, China in 2009, the M.S. degree in Mechanical Engineering from Purdue University in 2011, and M.S. degree in Statistics in 2014 and Ph.D. degree in Industrial and Systems Engineering in 2015, both from University of Wisconsin-Madison.
Currently, he is an Assistant Professor in the Dept. of Industrial Engineering and Management at Peking University, Beijing, China. He was an Assistant Professor at the Dept. of IMSE at UTEP, TX, USA from 2015 to 2017.

His research interests are mainly in quality control and reliability engineering of intelligent manufacturing and complex systems through engineering-informed machine learning and advanced data analytics. He is a recipient of the STARS
Award from the University of Texas Systems, Overseas Distinguished Young Scholars from China, P$\&$G Faculty Fellowship, BOSS Award from MSEC, and several Best Paper Award/Finalists from INFORMS/IISE Annual Meeting.  He is an Associate Editor of the Journal of Intelligent Manufacturing, and a member of IEEE, INFORMS, IISE, and SME.
\end{IEEEbiography}

\appendices
\section{Derivation of covariance function in convolution process}
\label{appendix: covariance derivation}
\color{black}
For the convolution process:
\begin{align}
f_i(\bm{x})=g_i(\bm{x})\ast Z (\bm{x})=\int_{-\infty}^{\infty}g_i (\bm{x}-\bm{u}) Z (\bm{u}) d\bm{u}, \notag
\end{align}
If $Z(\bm{x})$ is a commonly used white Gaussian noise process, i.e., ${\rm cov}\left(Z(\bm{x}), Z(\bm{x}^{\prime})\right)=\delta(\bm{x}-\bm{x}^{\prime})$ and $\mathbb{E}(Z(\bm{x}))=0$, then the cross covariance is derived as:
\begin{align}
&{\rm cov}_{ij}^f \left(\bm{x}, \bm{x}^{\prime} \right) ={\rm cov}\{ g_i(\bm{x})\ast Z (\bm{x}), g_j(\bm{x}^{\prime})\ast Z (\bm{x}^{\prime})\} \notag\\
&=\mathbb{E} \left\{  \int_{-\infty}^{\infty}g_i (\bm{x}-\bm{u}) Z (\bm{u}) d\bm{u} \int_{-\infty}^{\infty}g_j (\bm{x}^{\prime}-\bm{u}^{\prime}) Z (\bm{u}^{\prime}) d\bm{u}^{\prime} \right\} \notag\\
&=\int_{-\infty}^{\infty}\int_{-\infty}^{\infty}g_i (\bm{u})g_j (\bm{u}^{\prime}) \mathbb{E} \left\{ Z (\bm{x}-\bm{u})  Z (\bm{x}^{\prime}-\bm{u}^{\prime}) \right\} d\bm{u}d\bm{u}^{\prime} \notag\\
&=\int_{-\infty}^{\infty} \int_{-\infty}^{\infty} g_i(\bm{u})g_j(\bm{u}^{\prime}) \delta(\bm{x}-\bm{u}-\bm{x}^{\prime}+\bm{u}^{\prime}) d \bm{u}d\bm{u}^{\prime}  \notag\\
&=\int_{-\infty}^{\infty} g_i(\bm{u})g_j(\bm{u}-\bm{v})d \bm{u},
\label{eq:cov in convolution process-derivation}
\end{align}
where $\bm{v}=\bm{x}-\bm{x}^{\prime}$ and the last equality is based on the property of Dirac function that $\int g(\bm{u}^{\prime})\delta(\bm{u}^{\prime}-\bm{x})d\bm{u}^{\prime}=g(\bm{x})$.

For our MGCP structure:
\begin{align}
y_i(\bm{x})&=f_i(\bm{x})+\epsilon_i(\bm{x})=g_{ii}(\bm{x})\ast Z_i(\bm{x})+\epsilon_i(\bm{x}), i \in \mathcal{I}^S \notag\\
y_t(\bm{x})&=f_t(\bm{x})+\epsilon_t(\bm{x})=\sum_{j \in \mathcal{I}}g_{jt}(\bm{x}) \ast Z_j(\bm{x})+\epsilon_t(\bm{x}), \notag
\end{align}
the source-target covariance function can be calculated as:
\begin{align}
{\rm cov}_{it}^f(\bm{x},\bm{x}^{\prime}) 
&={\rm cov} (f_i(\bm{x}),f_t(\bm{x}^{\prime})) \notag\\
&={\rm cov}\left\{g_{ii}(\bm{x})\ast Z_i(\bm{x}), \sum_{j \in \mathcal{I}}g_{jt}(\bm{x}^{\prime}) \ast Z_j(\bm{x}^{\prime}) \right\}\notag\\
&= \sum_{j \in \mathcal{I}} {\rm cov}\left\{ g_{ii}(\bm{x})\ast Z_i(\bm{x}), g_{jt}(\bm{x}^{\prime}) \ast Z_j(\bm{x}^{\prime}) \right\} \notag\\
&=\int_{-\infty}^{\infty} g_{ii}(\bm{u})g_{it}(\bm{u}-\bm{v})d\bm{u},\quad i \in \mathcal{I}^S 
\end{align}
where the last equality is based on \Eqref{eq:cov in convolution process}, and $\bm{v}=\bm{x}-\bm{x}^{\prime}$. In the same way, we can derive the auto-covariance as
\begin{align}
{\rm cov}_{ii}^f(\bm{x},\bm{x}^{\prime})&=\int_{-\infty}^{\infty} g_{ii}(\bm{u})g_{ii}(\bm{u}-\bm{v})d\bm{u}, i\in \mathcal{I}^S \notag\\
{\rm cov}_{tt}^f(\bm{x},\bm{x}^{\prime})&=\sum_{j \in \mathcal{I}}\int_{-\infty}^{\infty} g_{jj}(\bm{u})g_{jt}(\bm{u}-\bm{v})d\bm{u}.\notag
\end{align}
\color{black}

\section{Proof of Theorem 1}
\label{appendix: conditional distribution}
Suppose that $g_{it}(\bm{x})=0, \forall i \in  \mathcal{U} \subseteq \mathcal{I}^S$ for all $\bm{x}\in \mathcal{X}$. 
For notational convenience, suppose $\mathcal{U}=\{1,2,...,h|h\leq q\}$, then the predictive distribution of the model at any new input $\bm{x}_*$ is unrelated with $\{f_1,f_2,...,f_h\}$ and is reduced to:
\begin{align}
p(y_t(\bm{x}_{*}) | \bm{y})=\mathcal{N}(&\bm{k}_{+}^T \bm{C}_{+}^{-1} \bm{y}_{+}, \notag\\
&{\rm cov}_{tt}^f(\bm{x}_{*},\bm{x}_{*})+\sigma_t^2-\bm{k}_{+}^T \bm{C}_{+}^{-1} \bm{k}_{+}), \nonumber
\end{align}
where $\bm{k}_{+}=(\bm{K}_{h+1,*}^T,...,\bm{K}_{q,*}^T,\bm{K}_{t,*}^T)^T$, $\bm{y}_{+}=(\bm{y}_{h+1}^T,...,\bm{y}_{q}^T,\bm{y}_{t}^T)^T$, and 
$$\bm{C}_{+}=
\renewcommand{\arraystretch}{1.2}
\begin{pmatrix}
\bm{C}_{h+1,h+1} & \cdots & \bm{0} & \bm{C}_{h+1,t} \\
\vdots & \ddots & \vdots & \vdots  \\
\bm{0}  & \cdots & \bm{C}_{q,q}  & \bm{C}_{q,t} \\
\bm{C}_{h+1,t}^T & \cdots & \bm{C}_{q,t}^T & \bm{C}_{t,t}
\end{pmatrix}.$$

\textbf{Proof.} Recall that 
\begin{align}
{\rm cov}_{jt}^y(\bm{x},\bm{x}^{\prime})&={\rm cov}_{jt}^f(\bm{x},\bm{x}^{\prime}) \notag\\
&=\int_{-\infty}^{\infty} g_{jj}(\bm{u})g_{jt}(\bm{u}-\bm{v})d\bm{u}, \notag\\
{\rm cov}_{tt}^y(\bm{x},\bm{x}^{\prime})&={\rm cov}_{tt}^f(\bm{x},\bm{x}^{\prime})+\sigma_t^2 \delta(\bm{x}-\bm{x}^{\prime}) \notag\\
&=\sum_{h \in \mathcal{I}}\int_{-\infty}^{\infty} g_{hh}(\bm{u})g_{ht}(\bm{u}-\bm{v})d\bm{u}+\sigma_t^2 \delta(\bm{x}-\bm{x}^{\prime}),  \nonumber
\end{align}
for all $j \in \{1,2,...,q\}$, so $g_{it}(\bm{x})=0,i\in \{1,2,...,h| h \leq q\}$ implies that ${\rm cov}_{it}^y \left(\bm{x},\bm{x}^{\prime} \right)=0$ for all $i \in \{1,2,...,h\}$ and 
$${\rm cov}_{tt}^y(\bm{x},\bm{x}^{\prime})=\sum_{i=h+1}^t \int_{-\infty}^{\infty} g_{ii}(\bm{u})g_{it}(\bm{u}-\bm{v})d\bm{u}+\sigma_t^2 \delta(\bm{x}-\bm{x}^{\prime}).$$
Therefore, we have that $\bm{C}_{i,t}=0, i \in \{1,2,...,h\}$ and partition covariance matrix 
$\bm{C}=
\begin{pmatrix}
\bm{C}_{-} & \bm{0} \\
\bm{0} & \bm{C}_{+}
\end{pmatrix}$, 
where
$\bm{C}_{-}=
\begin{pmatrix}
\bm{C}_{1,1} & \bm{0} & \cdots & \bm{0}  \\
\bm{0} & \bm{C}_{2,2} & \cdots & \bm{0} \\
\vdots & \vdots & \ddots & \vdots \\
\bm{0} & \bm{0} & \cdots & \bm{C}_{h,h} \\
\end{pmatrix}$.

The predictive distribution at point $\bm{x}_{*}$ is 
$$y_t(\bm{x}_{*})\sim N(\bm{K}_*^T \bm{C}^{-1} \bm{y}, {\rm cov}_{tt}^f (\bm{x}_{*},\bm{x}_{*})+\sigma_t^2-\bm{K}_*^T \bm{C}^{-1} \bm{K}_*).$$
Also, based on that ${\rm cov}_{it}^y(\bm{x},\bm{x}^{\prime})=0$ for all $i \in \{1,2,...,h\}$, we have that $\bm{K}_*=\left(\bm{0}, \bm{k}_{+}^T\right)^T$. Let $\bm{y}_{-}=\left(\bm{y}_{1}^T,...,\bm{y}_{h}^T\right)^T$, then $\bm{y}=\left(\bm{y}_{-}^T,\bm{y}_{+}^T\right)^T$. Therefore, 
\begin{align}
\bm{K}_*^T \bm{C}^{-1} \bm{y} &=(\bm{0}, \bm{k}_{+}^T)
\begin{pmatrix}
\bm{C}_{-} & \bm{0} \\
\bm{0} & \bm{C}_{+}
\end{pmatrix}^{-1}
(\bm{y}_{-}^T,\bm{y}_{+}^T)^T  \notag\\
&=(\bm{0}, \bm{k}_{+}^T)
\begin{pmatrix}
\bm{C}_{-}^{-1} & \bm{0} \notag\\
\bm{0} & \bm{C}_{+}^{-1}
\end{pmatrix}
(\bm{y}_{-}^T,\bm{y}_{+}^T)^T \notag\\
&=\bm{k}_{+}^T \bm{C}_{+}^{-1} \bm{y}_{+}, \notag\\
\bm{K}_*^T \bm{C}^{-1} \bm{K}_* &=(\bm{0}, \bm{k}_{+}^T)
\begin{pmatrix}
\bm{C}_{-} & \bm{0} \\
\bm{0} & \bm{C}_{+}
\end{pmatrix}^{-1}
(\bm{0}, \bm{k}_{+}^T)^T  \notag\\
&=\bm{k}_{+}^T \bm{C}_{+}^{-1} \bm{k}_{+}. \nonumber
\end{align}

Note that the auto-covariance matrix of target output $f_t$, $\bm{C}_{tt}$, is also unrelated with observed data $\{\bm{X}_i | i=1,2,...,h\}$ which from source output $\{f_i| i=1,2,...,h\}$. As a result, the predictive distribution is totally independent on these outputs. Proof completes.

\section{Regularity conditions}
\label{appendix:regularity conditions}
In this part, we state the regularity conditions for the consistency theorem of the MLE $\hat{\bm{\theta}}_{\#}$, which are formulated in \cite{Basawa2014}.

Denote $\bm{y}$ with total $N$ observations as $\bm{y}^{N}$, and let 
$$p_k(\bm{\theta}) =\frac{ p(\bm{y}^{k}| \bm{\theta}) } {p(\bm{y}^{k-1}| \bm{\theta}) }$$
for each $k$. Assume $p_k(\bm{\theta})$ is twice differentiable with respect to $\bm{\theta}$ in a neighborhood of $\bm{\theta}^*$. Also assume that the support of $p(\bm{y}^{N}| \bm{\theta})$ is independent of $\bm{\theta}$ in the neighborhood. Define $\phi_k(\bm{\theta}) = \log p_k(\bm{\theta})$, and its first derivative $\phi_k^{\prime}(\bm{\theta})$, second derivative $\phi_k^{\prime\prime}(\bm{\theta})$. 

For simplicity and without loss of generality, we only consider the conditions for one-dimensional case. Define $\phi_k^{*\prime}=\phi_k^{\prime}({\theta}^*)$ and $\phi_k^{*\prime\prime}=\phi_k^{\prime\prime}({\theta}^*)$. Let $\mathcal{F}_N$ be the $\sigma$-field generated by $y_j, 1 \leq j \leq N$, and $\mathcal{F}_0$ be the trivial $\sigma$-field. Define the random variable $i_k^* = var(\phi_k^{*\prime} | \mathcal{F}_{k-1})=\mathbb{E}[(\phi_k^{*\prime})^2 | \mathcal{F}_{k-1}]$ and $I_N^* = \sum_{k=1}^N i_k^*$. Define $S_N = \sum_{k=1}^N \phi_k^{*\prime}$ and $S_N^* = \sum_{k=1}^N \phi_k^{*\prime\prime}+I_N^*$. If the following conditions hold:
\begin{itemize}
\item[(c1)] $\phi_k({\theta})$  is thrice differentiable in the neighborhood of ${\theta}^*$. Let $\phi_k^{*\prime\prime\prime}=\phi_k^{\prime\prime\prime}({\theta}^*)$ be the third derivative,
\item[(c2)] Twice differentiation of $\int p(\bm{y}^N | \theta) d \mu^N(\bm{y}^N)$ with respect to $\theta$ of exists in the neighborhood of ${\theta}^*$,
\item[(c3)] $\mathbb{E}|\phi_k^{*\prime\prime}| < \infty$ and $\mathbb{E}|\phi_k^{*\prime\prime} +  (\phi_k^{*\prime})^2| < \infty$.
\item[(c4)] There exists a sequence of constants $K(N) \rightarrow \infty$ as $N \rightarrow \infty$ such that:
			\begin{itemize}
			\item[(i)] $K(N)^{-1} S_N \overset{p}{\rightarrow} 0$,
			\item[(ii)] $K(N)^{-1} S_N^* \overset{p}{\rightarrow} 0$,
			\item[(iii)] there exists $a(\theta^*) > 0$ such that $\forall \epsilon > 0$,  $P[K(N)^{-1}I_N^* \geq 2a(\theta^*)] \geq 1-\epsilon$ for all $N \geq N(\epsilon)$,
			\item[(iv)] $K(N)^{-1} \sum_{k=1}^N \mathbb{E}|\phi_k^{*\prime\prime\prime}| < M < \infty$ for all $N$,
			\end{itemize}
\end{itemize}
then the MLE $\hat{{\theta}}_{\#}$ is consistent for $\theta^*$. There exists a sequence $r_N$ such that $r_N \rightarrow \infty$ as $N\rightarrow \infty$, i.e.,
\begin{align*}
\| \hat{{\theta}}_{\#}-{\theta}^*\| = O_P(r_N^{-1}).
\end{align*} 


\section{Proof of Theorem 2}
\label{appendix:consistency theorem}
Suppose that the MLE for $L(\bm{\theta}|\bm{y})$, $\hat{\bm{\theta}}_{\#}$, is $r_N$ consistent, i.e., satisfying \Eqref{non-penalized maximum log-likelihood estimator consistency}. If $\max \{ |\mathbb{P}^{\prime \prime}_{\gamma}({\theta}_{i0}^*)|: {\theta}_{i0}^* \neq 0\} \rightarrow 0$, then there exists a local maximizer $\hat{\bm{\theta}}$ of  $L_{\mathbb{P}}(\bm{\theta}|\bm{y})$ s.t. $\|\hat{\bm{\theta}}-\bm{\theta}^{*} \|=O_P(r_N^{-1}+r_0)$, where $r_0=\max \{ |\mathbb{P}^{\prime}_{\gamma}({\theta}_{i0}^*)|: {\theta}_{i0}^* \neq 0\}$.

\textbf{Proof.} Recall the assumptions in Section \ref{sec:3-statistical}. For the unpenalized log-likelihood $L(\bm{\theta})$, the MLE $\hat{\bm{\theta}}_{\#}$ is $r_N$ consistent where $r_N$ is a sequence such that $r_N \rightarrow \infty$ as $N \rightarrow \infty$. And we have that $L^{\prime}(\bm{\theta}^{*})=O_P(r_N)$ and $\bm{I}_N(\bm{\theta}^*)=O_P(r_N^{2})$, which are the standard argument based on the consistency of estimator. Based on that, we aim to study the asymptotic properties of the penalized likelihood $L_{\mathbb{P}}(\bm{\theta})=L(\bm{\theta})-r_N^2\mathbb{P}_{\gamma}(\bm{\theta}_0)$. Here we multiply the penalty function by $r_N^2$ to avoid that penalty term degenerates as $N \rightarrow \infty$. The following proof is similar to that of Fan and Li \cite{Fan2001} but based on dependent observations.

To prove theorem 2, we need to show that for any given $\epsilon>0$, there exists a large constant  $U$ such that:
\begin{align}
P\left\{ \underset{\|\bm{u}\|=U} \sup L_{\mathbb{P}}(\bm{\theta}^{*}+r_N^+\bm{u})<L_{\mathbb{P}}(\bm{\theta}^{*}) \right\} \geq 1-\epsilon,
\label{t2-proof-target}
\end{align}
where $r_N^+=r_N^{-1}+r_0$. This implies that with probability at least $1-\epsilon$ there exists a local maximum in the ball $\{\bm{\theta}^{*}+r_N^{+}\bm{u}: \| \bm{u} \| \leq U\}$. So the local maximizer $\hat{\bm{\theta}}$ satisfies that $\| \hat{\bm{\theta}}-\bm{\theta}^{*} \|=O_P(r_N^{+})$.

By $\mathbb{P}_{\gamma}(0)=0$, we have
\begin{align}
&L_{\mathbb{P}}(\bm{\theta}^{*}+r_N^+\bm{u})-L_{\mathbb{P}}(\bm{\theta}^{*}) \notag\\
&\leq L(\bm{\theta}^{*}+r_N^+\bm{u})-L(\bm{\theta}^{*}) \notag\\
&\quad -r_N^2\sum_{i=h+1}^q \left[ \mathbb{P}_{\gamma}(|{\theta}_{i0}^{*}+r_N^{+}u_{i0}|)-\mathbb{P}_{\gamma}(|{\theta}_{i0}^{*}|) \right], \nonumber
\end{align}
where $h$ and $q$ are the number of zero components and all components in ${\theta}_{i0}^{*}$, and $u_{i0}$ is the element corresponding to ${\theta}_{i0}$ in $\bm{u}$. Let $\bm{I}_N(\bm{\theta}^{*})$ be the finite and positive definite information matrix at $\bm{\theta}^{*}$ with $N$ observations. Applying a Taylor expansion on the likelihood function, we have that 
\begin{align}
&L_{\mathbb{P}} (\bm{\theta}^{*}+r_N^+\bm{u})-L_{\mathbb{P}}(\bm{\theta}^{*}) \notag\\
& \leq r_N^{+} L^{\prime}(\bm{\theta}^{*})^{T}\bm{u}-\frac{1}{2}(r_N^{+})^2\bm{u}^T \bm{I}_N(\bm{\theta}^{*})\bm{u}[1+o_P(1)] \notag\\
&\quad -r_N^2\sum_{i=h+1}^q \Big\{ r_N^+ \mathbb{P}_{\gamma}^{\prime}(|{\theta}_{i0}^{*}|) {\rm sign} ({\theta}_{i0}^{*}) u_{i0}  \notag\\
&\qquad \qquad \qquad +\frac{1}{2}(r_N^+)^2\mathbb{P}_{\gamma}^{\prime \prime}(|{\theta}_{i0}^{*}|)u_{i0}^2[1+o_P(1)] \Big\},
\label {t2-proof-Taylor}
\end{align}
Note that $\|L^{\prime}(\bm{\theta}^{*})\|=O_P(r_N)$ and $\bm{I}_N(\bm{\theta}^{*})=O_P(r_N^2)$. so the first term on the right-hand side of \Eqref{t2-proof-Taylor} is on the order $O_P(r_N^+r_N)$, while the second term is $ O_P\left( (r_N^+r_N)^2 \right) $. By choosing a sufficient large $U$, the first term can be dominated by the second term uniformly in $\|u\|=U$. Besides, the absolute value of the third term is bounded by
$$\sqrt{q-h}r_N^2r_N^+r_0 \|\bm{u}\|+(r_Nr_N^+)^2 \max \{ |\mathbb{P}^{\prime \prime}_{\gamma}({\theta}_{i0})|: {\theta}_{i0} \neq 0\} \|\bm{u}\|^2,$$
which is also dominated by second term as it is on the order of $o_P\left((r_Nr_N^+)^2 \right)$. Thus, \Eqref{t2-proof-target} holds and the proof completes.

\section{Proof of Theorem 3}
\label{appendix:sparsity theorem}
Let $\bm{\theta}_{10}^{*}$ and $\bm{\theta}_{20}^{*}$ contain the zero and non-zero components in  $\bm{\theta}_{0}^{*}$ respectively.
Assume the conditions in \autoref{theorem:consistency} also hold, and $\hat{\bm{\theta}}$ is $r_N$ consistent by choosing proper $\gamma$ in $\mathbb{P}_{\gamma}(\bm{\theta}_0)$. If 
$\underset{N \rightarrow \infty}{\lim \inf}\ \underset{\theta \rightarrow 0^+}{\lim \inf}\ \gamma^{-1}\mathbb{P}^{\prime}_{\gamma}(\theta) >0$ and $(r_N \gamma)^{-1} \rightarrow 0$, 
then $$\underset{N \rightarrow \infty}{\lim}P \left( \hat{\bm{\theta}}_{10}=\bm{0} \right)=1.$$

\textbf{Proof.} To prove this theorem, we only need to prove that for a small $\epsilon_N=Ur_N$, where $U$ is a given constant and $i=1,...,s$,
\begin{align}
\frac{\partial L_{\mathbb{P}} (\bm{\theta})}{\partial \theta_{i0}}\theta_{i0}<0, 0<|\theta_{i0}|<\epsilon_N.
\label{t3-proof-target}
\end{align}
By Taylor's expansion,
\begin{align}
&\frac{\partial L_{\mathbb{P}} (\bm{\theta})}{\partial \theta_{i0}} 
=\frac{\partial L (\bm{\theta})}{\partial \theta_{i0}}-r_N^2 \mathbb{P}_{\gamma}^{\prime}(|\theta_{i0}|) {\rm sign} (\theta_{i0}) \notag\\
&=\frac{\partial L (\bm{\theta}^{*})}{\partial \theta_{i0}}+\left[ \partial \left( \frac{\partial L (\bm{\theta}^{*})}{\partial \theta_{i0}} \right)/ \partial \bm{\theta}\right]^T(\bm{\theta}-\bm{\theta}^{*})[1+o_P(1)]\notag\\
&\quad -r_N^2\mathbb{P}_{\gamma}^{\prime}(|\theta_{i0}|) {\rm sign} (\theta_{i0}). \nonumber
\end{align}
As $\frac{\partial L (\bm{\theta})}{\partial \theta_{i0}}=O_P(r_N)$, $\partial \left( \frac{\partial L (\bm{\theta}^{*})}{\partial \theta_{i0}} \right)/ \partial \theta_j =O_P(r_N^2)$ by the standard argument for $r_N$ consistent estimator, thus
\begin{align}
\frac{\partial L_{\mathbb{P}} (\bm{\theta})}{\partial \theta_{i0}}&=O_P(r_N)-r_N^2\mathbb{P}_{\gamma}^{\prime}(|\theta_{i0}|) {\rm sign} (\theta_{i0}) \notag\\
&=r_N^2\gamma \left( O_P(\frac{1}{r_N\gamma})-\gamma^{-1}\mathbb{P}_{\gamma}^{\prime}(|\theta_{i0}|) {\rm sign} (\theta_{i0}) \right). \nonumber
\end{align}
Because that $\underset{N \rightarrow \infty}{\lim \inf}\ \underset{\theta \rightarrow 0^+}{\lim \inf}\ \gamma^{-1}\mathbb{P}^{\prime}_{\gamma}(\theta) >0$ and $(r_N \gamma)^{-1} \rightarrow 0$, $\frac{\partial L_{\mathbb{P}} (\bm{\theta})}{\partial \theta_{i0}}$ will be positive while $\theta_{i0}$ is negative and vise versa. As a result, \Eqref{t3-proof-target} follows. Proof completes.

\section{Interpretation of the benchmark: MGCP-RF}
\label{appendix:MGCP-RF}
\begin{figure}[H]
\centering
\includegraphics[width=2.5in]{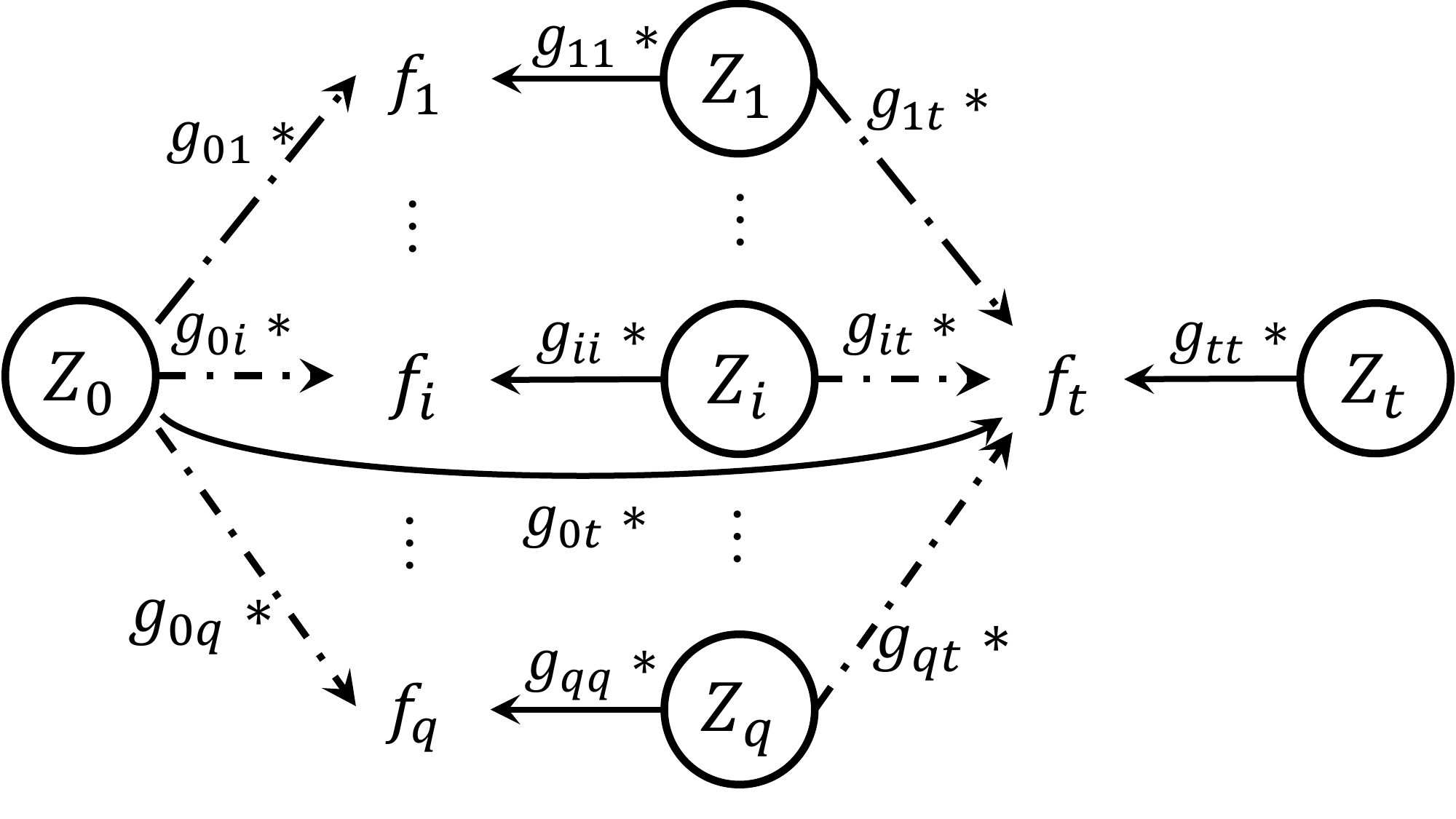}
\caption{The structure of MGCP-RF}
\label{fig:structure-MGCP-RF}
\end{figure}

\color{black}
The illustration of MGCP-RF is shown in \Figref{fig:structure-MGCP-RF}.

In this structure, target $f_t$ is generated by three kinds of latent process: $Z_0(\bm{x})$, $\{Z_i (\bm{x})\}_{i=1}^q$ and $Z_t(\bm{x})$. As $Z_0(\bm{x})$ is the common process shared by sources, the covariance matrix blocks between source $f_i$ and the other outputs are zero only when the scale parameters in $g_{0i}(\bm{x})$ and $g_{it}(\bm{x})$  are zero simultaneously. Thus, the marginalized covariance matrix $\bm{C}_{+}$ in Theorem 1 will be:
$$\bm{C}_{+}=
\renewcommand{\arraystretch}{1.2}
\begin{pmatrix}
\bm{C}_{h+1,h+1} & \cdots & \bm{C}_{h+1,q} & \bm{C}_{h+1,t} \\
\vdots & \ddots & \vdots & \vdots  \\
\bm{C}_{h+1,q}^T  & \cdots & \bm{C}_{q,q}  & \bm{C}_{q,t} \\
\bm{C}_{h+1,t}^T & \cdots & \bm{C}_{q,t}^T & \bm{C}_{t,t}
\end{pmatrix}.$$
The difference to MGCP-R is that covariance among the remaining sources $\{f_i\}_{i=h+1}^q$ can be modeled. This structure is indeed more comprehensive but with the cost of a half more parameters than MGCP-R. The cost will increase if we use more latent process to model the correlation among sources.

To realize the effect of shrinking $g_{0i}(\bm{x})$ and $g_{it}(\bm{x})$ at the same time, group-L1 penalty is used and the penalized log-likelihood function is:
\begin{align}
\underset{\bm{\theta}}{\rm max}\ L_{\mathbb{P}}(\bm{\theta}| \bm{y})=&L(\bm{\theta}| \bm{y})-\gamma \sum_{i=1}^q \sqrt{\alpha_{0i}^2+\alpha_{it}^2}, \notag
\label{eq:regularized log-likelihood with group L1}
\end{align}
\color{black}

\section{Influence of tuning-parameter}
\label{appendix:tuning-parameter}
\begin{figure}[H]
\centering
\includegraphics[width=3.0in]{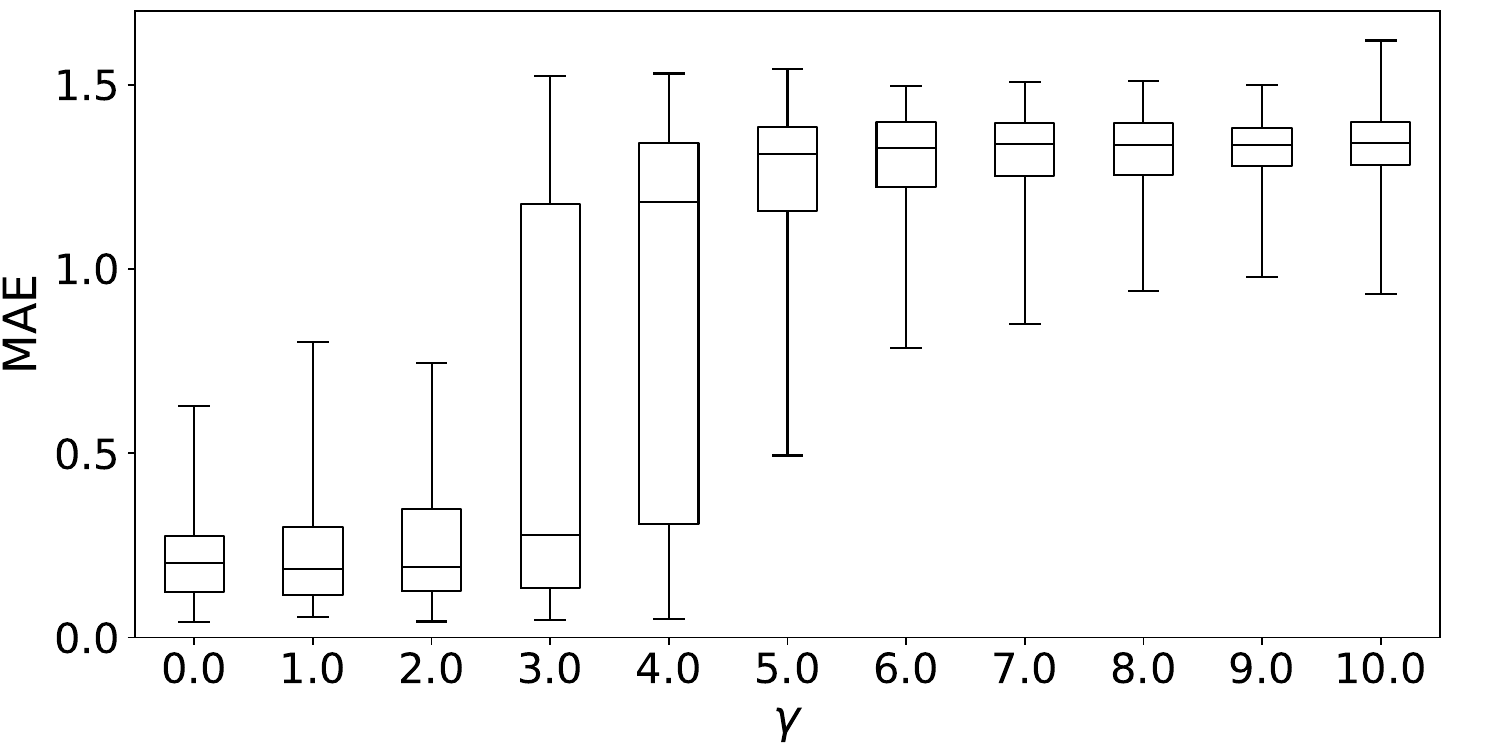}
\caption{Prediction error with different $\gamma$ in 100 repetition.}
\label{fig:MAE with gamma}
\end{figure}

\begin{figure}[H]
\centering
\includegraphics[width=3.0in]{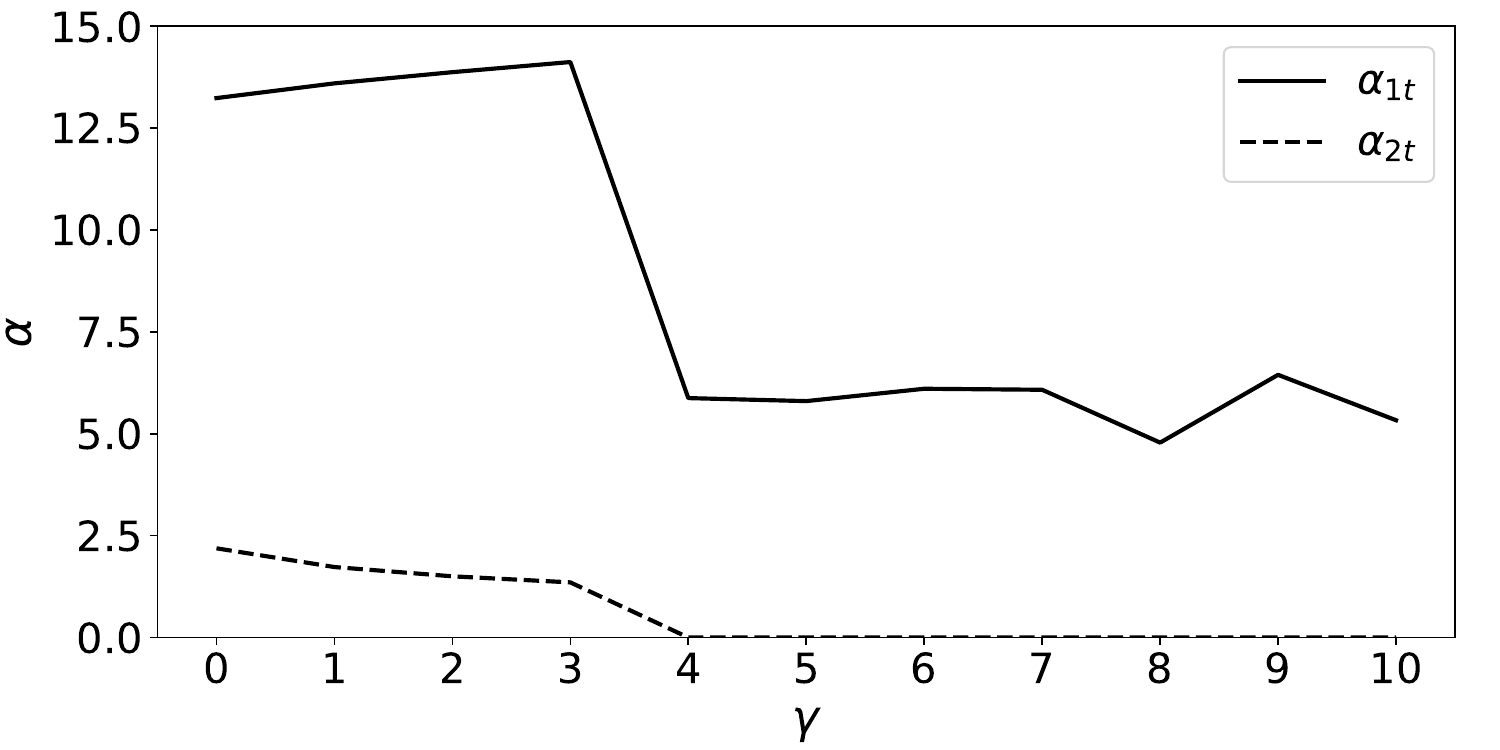}
\caption{Estimated values of $\alpha_{1t}, \alpha_{2t}$ in one repetition.}
\label{fig:alpha with gamma}
\end{figure}

\color{black}
To test the influence of the tuning-parameter $\gamma$ in our model, we conduct the following experiment. Based on the same dataset in the 1D example of simulation case I, we construct MGCP-R model only with sources $f_1$ and $f_2$, and let $\gamma$ vary from $0$ to $10$ at a step of 1. Note that MGCP-T is equal to the model with $\gamma=0$. The boxplot of MAE with respect to different values of $\gamma$ is shown in Fig. \ref{fig:MAE with gamma}. The estimated value of $\alpha_{1t}, \alpha_{2t}$ in one repetition is presented in Fig. \ref{fig:alpha with gamma}. It can be seen that as $\gamma$ increases, source $f_2$ will be excluded from the prediction of target, leading to an increased prediction error. In practice, cross-validation can be used to select an optimal tuning-parameter.

\color{black}

\end{document}